\pdfoutput=1
\documentclass{article}
\usepackage{graphicx}
\usepackage{amsmath}
\usepackage{amsthm}

\usepackage{algorithm}
\usepackage{algorithmic}
\usepackage{bbm}
\usepackage[round]{natbib}
\usepackage{url}
\usepackage{caption}
\usepackage{subcaption}
\usepackage{fullpage}

\begin{document}

\title{{Perturbed Gibbs Samplers for Synthetic Data Release}}

\author{Yubin Park and Joydeep Ghosh}
\date{Department of Electrical and Computer Engineering\\ The University of Texas at Austin, USA}

\maketitle

\begin{abstract}
We propose a categorical data synthesizer with a quantifiable disclosure risk. 
Our algorithm, named Perturbed Gibbs Sampler, can handle high-dimensional categorical data that are often intractable to represent as contingency tables.
The algorithm extends a multiple imputation strategy for fully synthetic data by utilizing feature hashing and non-parametric distribution approximations.
California Patient Discharge data are used to demonstrate statistical properties of the proposed synthesizing methodology.
Marginal and conditional distributions, as well as the coefficients of regression models built on the synthesized data are compared to those obtained from the original data.
Intruder scenarios are simulated to evaluate disclosure risks of the synthesized data from multiple angles.
Limitations and extensions of the proposed algorithm are also discussed.
\end{abstract}

\section{Introduction}

Public use data, which are often released by government and other data collecting agencies, typically need to to satisfy two competing objectives: maintaining relevant statistical properties of the original data and protecting privacy of individuals.
To address these two goals, various statistical disclosure limitation techniques have been developed \citep{Willenborg2001}.
Some popular disclosure techniques are data swapping \citep{Dalenius1978,Fienberg2005}, top-coding, feature generalization such as $k$-anonymity \citep{Sweeney2002} or $l$-diversity \citep{Machanavajjhala2007}, and additive random noise with measurement error models \citep{Fuller1993}.
Each method has distinct utility and risk aspects.
In practice, a disclosure limitation technique is carefully chosen by domain experts and statisticians. 
Sometimes, multiple techniques are mixed and applied to a single dataset to achieve better privacy protection before being released to the public \citep{web:cms}.
Such public use datasets have served as valuable information sources for decision makings in economics, healthcare, and business analytics.

The generation of synthetic data, proposed by Rubin (\citeyear{Rubin1993}), is an alternative approach to data-transforming disclosure techniques. 
Multiple imputation, which was originally developed to impute missing values in survey responses \citep{Rubin1987}, is used to generate either partially or fully synthetic data.
As synthetic data preserves the structure and resolution of the original data, preprocessing steps and analytical procedures on synthetic data can be effortlessly transferred to the original data.
This aspect has contributed to popular adoption of synthetic data in diverse research areas.
Thus far, there has been notable progress on valid inferences using synthetic data and extensions to different applications:
Abowd and Woodcock (\citeyear{Abowd2001}) synthesized a French longitudinal linked database, and Raghunathan et al. (\citeyear{Raghunathan2003}) provided general methods for obtaining valid inferences using multiply imputed data.
Beyond typically used generalized linear models, decision trees models, such as CART and Random forests, can also be used as imputation models in multiple imputation \citep{Reiter2003,Caiola2010}.
Some illustrative empirical studies have used U.S. census data \citep{Drechsler2010}, German business database \citep{Reiter2010}, and U.S. American Community Survey \citep{Sakshaug2011}.

A very different approach to imputing missing values in binary or binarized datasets can be taken using association rule mining.
Vreenken and Siebes (\citeyear{Vreeken2008}) used the minimum description length principle to develop a set of heuristics that are used to (approximately) represent the dataset in terms of a concise set of frequent itemsets. 
These rules can be then used to impute missing values, and in principle could also be used to generate synthetic data. 
However, their current work does not quantify the privacy afforded by the compressed data or use a given privacy criterion to determine the derived itemsets.

The two competing requirements for public use data similarly apply to synthetic data disclosure.
Synthetic data need to be accurate enough to answer relevant statistical queries without revealing private information to third parties.
Statistical properties of synthetic data are primarily determined by imputation models \citep{Reiter2005}, and models that are too accurate tend to leak private information \citep{Abowd2008}.

The balance between accuracy and privacy can also be addressed by using cryptographic privacy measures such as $\epsilon$-differential privacy \citep{Dwork2006:icalp}. 
However, several attempts to achieve such strong privacy guarantees have shown to be impractical to implement.
For example, Barak et al. (\citeyear{Barak2007}) showed that it is possible to release contingency tables under the differential privacy regime using Fourier transform and additive Laplace noise.  
However, this proposed release mechanism was later criticized for being too conservative and disrupting statistical properties of the original data \citep{Yang2012,Charest2012}.
On the other hand, Soria-Cormas and Drechsler (\citeyear{SoriaCormas2013}) claimed that $\epsilon$-differential privacy can be a useful privacy measure when disclosing a large size of data with a limited number of variables.  
For example, differentially private synthetic data have been demonstrated using the Census Bureau's OnTheMap data that consists of approximately one million records with two variables \citep{Machanavajjhala2008}. 

In this paper, we propose a \textit{practical} multi-dimensional categorical data synthesizer that satisfies $\epsilon$-differential privacy.
The proposed synthesizer can handle multi-dimensional data that are not practical to be represented as contingency tables.
We demonstrate our algorithm using a subset of California Patient Discharge data, and generate multiple synthetic discharge datasets.
Although $\epsilon$-differential privacy is extensively used in our algorithm analyses, we note that $\epsilon$-differential privacy is one of many descriptive measures for disclosure risks.
Differential privacy is a measure for functions, not for data \citep{Fienberg2010}, and this measure can be overly pessimistic for data-specific applications.
Thus, we also evaluate disclosure risks of the proposed algorithms using the population uniqueness of synthetic records \citep{Dale2001} and indirect-matching probabilistic disclosure risks \citep{Duncan1986}.
To measure the statistical similarities between synthetic and the original data, we compare marginal and conditional distributions, and regression coefficients from the synthesized data to those from the original data.

\begin{table}\caption{Synthesizer algorithms discussed in this paper. }\label{tab:models}
\begin{center}
\begin{small}
\begin{tabular}{| c | c | c |}
\hline
Name & Abbreviated Model Equation & Model Parameters \\ \hline
Contingency table & $\text{Pr}_{\mathcal{D}}(\mathbf{x})$ & non-parametric \\
Marginal Bayesian Bootstrap & $\prod_{i}^{M} \text{Pr}_{\mathcal{D}}({x_i})$ & non-parametric \\
Multiple imputation & $\prod_{i}^{M} \text{Pr}_{\hat{\mathbf{w}}} ({x_i} \mid \mathbf{x}_{-i} )$ & $\hat{\mathbf{w}}$: model coefficients for glm \\
Perturbed Gibbs Sampler (PeGS) & $\prod_{i}^{M} \text{Pr}_{\mathcal{D},\alpha}({x_i} \mid h(\mathbf{x}_{-i}) )$ & $\alpha$: privacy parameter\\
Block PeGS with Reset & $\prod_{b}^{B} \prod_{i}^{M} \text{Pr}_{\mathcal{D},\alpha}({x_i} \mid h(\mathbf{x}_{-i}) )$ & $B$: sample block size \\\hline
\end{tabular}
\end{small}
\end{center}
\end{table}

There are two brute-force methods to generate synthetic categorical data. 
As statistical properties of categorical data are perfectly captured in contingency tables, in theory, a synthetic sample $\mathbf{x}$ can be drawn directly from an $M$-way full contingency table $\text{Pr}_{\mathcal{D}}(\mathbf{x})$, where $M$ is the total number of features.
For data with a small number of features, this contingency table can be estimated by either direct counting or log-linear models \citep{Winkler2003,Winkler2010}.
However, this strategy does not scale for multi-dimensional datasets.
As we will see in Section~\ref{sec:empirical},  our experiment dataset has 13 features and their possible feature combinations are approximately 2.6 trillion.
More importantly, sampling from an exact distribution may reveal too much detail about the original data, thus this is not a privacy-safe disclosure method. 
On the other extreme, one may model the joint distribution as a product of univariate marginal distributions.
Although this approach can easily achieve differential privacy \citep{McClure2012}, the synthetic data loses critical joint distributional information about the original data. 

The proposed algorithm generates \textit{realistic but not real} synthetic samples by calibrating a privacy parameter $\alpha$.
In addition, the exponentially number of cells in a contingency table is avoided by using multiple imputation and feature hashing $h(\mathbf{x}_{-i})$ as follows:
\begin{align} 
\text{for}~i~&\text{in}~1:M \nonumber\\
x_i &\sim \text{Pr}_{\mathcal{D},\alpha} (x_i \mid h(\mathbf{x}_{-i}) )  \label{eq:pegs}
\end{align}
where $\text{Pr}_{\mathcal{D},\alpha} ( x_i \mid h(\mathbf{x}_{-i}) )$ is the compressed and perturbed conditional distribution of the $i$th feature and $M$ is the total number of features.
The joint probability distribution is represented as $M$ conditional distributions.
Note that the conditional distribution in Equation~(\ref{eq:pegs}) is not exact. 
The full condition $\mathbf{x}_{-i}$ is compressed using a hash function $h(\mathbf{x}_{-i})$ and perturbed by a privacy parameter $\alpha$.
Ignoring these two additional components i.e. $h(\mathbf{x}_{-i})$ and $\alpha$, if the probability is modeled using generalized linear models, then the proposed algorithm is the same as a multiple imputation algorithm for fully synthetic data.
The proposed synthesizer is named as Perturbed Gibbs Sampler (PeGS).
This is because the proposed sampling procedure can iterate more than once unlike multiple imputation, and is similar to the Gibbs sampler. 
Table~\ref{tab:models} summarizes synthesizer models that are described in this paper.
More details on this list and privacy guarantees for both one iteration and multiple iterations are described in Section~\ref{sec:pegs}.

The rest of this paper is organized as follows:
In Section~\ref{sec:background}, we cover the basics of multiple imputation and $\epsilon$-differential privacy. 
In Section~\ref{sec:pegs}, the details of the PeGS algorithms are illustrated, and the privacy guarantees of the proposed algorithms are derived. 
We demonstrate our algorithms using California Patient Discharge dataset in Section~\ref{sec:empirical}.
Finally, we discuss the limitation of the proposed methods and future extensions in Section~\ref{sec:conclusion}.

\section{Background}\label{sec:background}

In this section, we overview multiple imputation, $\epsilon$-differential privacy, and $l$-diversity.
They are primary building blocks of our synthesizer algorithm. 
We start by describing the original multiple imputation method for missing values, then illustrate its application to generating fully synthetic data.
Next, we visit the definition of $\epsilon$-differential privacy and some approaches to implement differentially private algorithms.
The definition of $l$-diversity and its variant definition for synthetic data are illustrated.

\subsection{Multiple Imputation}

Multiple imputation was originally developed to impute missing values in survey responses \citep{Rubin1987}, and it was later applied to generate synthetic data. 
Let us start from the missing value imputation setting.
Consider a survey with two variables $x$ and $z$, $\mathcal{D} = \{(x,z)\}$, where some of the $x$ responses are missing.
Let $\mathcal{D}_{\text{obs}}$ be a subset of $\mathcal{D}$ where both $x$ and $z$ are observed.
The unobserved responses are imputed using samples from a posterior model as follows:
\begin{align*}
x \sim \text{Pr}_{\mathcal{D}_{\text{obs}}}(x \mid z)
\end{align*}
Note that  the posterior\footnote{Rubin (\citeyear{Rubin1987}) uses a different notation $\text{Pr}(x_{\text{nobs}} \mid x_{\text{obs}}, z)$, but they mean the same.} is modeled using the observed subset, and often obtained using generalized linear models or Bayesian Bootstraping methods \citep{Reiter2005}. 
To generate synthetic data, the process is repeated on the observed responses $x$ and $z$:
\begin{align*}
z \sim \text{Pr}_{\mathcal{D}_{\text{obs}}}(z \mid x )
\end{align*}
After this sampling step, a subset of fully synthetic responses is randomly sampled, and disclosed as public use data.
Typically, this entire process is repeated independently $K$ times to obtain $K$ different synthetic datasets.

Raghunathan et al. (\citeyear{Raghunathan2003}) showed that valid inferences can be obtained from multiply imputed synthetic data.
Let $Q$ be a function of $(x, z)$.
For example, $Q$ may represent the population mean of $(x, z)$ or the population regression coefficients of $x$ on $z$.
Let $q_i$ and $v_i$ be the estimate of $Q$ and its variance obtained from the $i$th synthetic dataset.
Then, valid inferences on $Q$ can be obtained as follows:
\begin{align*}
\bar{q}_K &= \sum_{i=1}^{K} q_i / K \\
T_s &= (1+\frac{1}{K})b_K - \bar{v}_K
\end{align*}
where $b_K = \sum_{i=1}^{K} (q_i - \bar{q}_K)^2 / (K-1)$ and $\bar{v}_K = \sum_{i=1}^{K} v_i /K$.
These two quantities $\bar{q}_K$ and $T_s$ estimate the original $Q$ and its variance.

\subsection{Differential Privacy}

Differential privacy \citep{Dwork2006:icalp} is a mathematical measure of privacy that quantifies disclosure risks of statistical functions.
To satisfy $\epsilon$-differential privacy, the inclusion or exclusion of any particular record in data cannot affect the outcome of functions by much.
Specifically, a randomized function $f: \mathcal{D} \rightarrow f(\mathcal{D})$ provides $\epsilon$-differential privacy, if it satisfies:
\begin{align*}
\frac{\text{Pr}(f(\mathcal{D}_1) \in \mathcal{S})}{\text{Pr}(f(\mathcal{D}_2) \in \mathcal{S})} \le \exp(\epsilon)
\end{align*}
for all possible $\mathcal{D}_1, \mathcal{D}_2 \in \mathcal{D}$ where $\mathcal{D}_1$ and $\mathcal{D}_2$ differ by at most one element, 
and $\forall \mathcal{S} \in \text{Range}(f(\mathcal{D}))$.
For a synthetic sample, this definition can be interpreted as follows \citep{McClure2012}:
\begin{align}
\frac{\text{Pr}_{\mathcal{D}_1} (\mathbf{x} )}{\text{Pr}_{\mathcal{D}_2} (\mathbf{x} )} \le \exp(\epsilon) \label{eq:diffpriv}
\end{align}
where $\mathbf{x}$ represents a random sample from synthesizers.
In other words, a data synthesizer $\text{Pr}_{\mathcal{D}}(\mathbf{x})$ is $\epsilon$-differentially private,
if the probabilities of generating $\mathbf{x}$ from $\mathcal{D}_1$ and $\mathcal{D}_2$ are indistinguishable to the extent of $\exp(\epsilon)$.

Several mechanisms have been developed to achieve differential privacy.
For numeric outputs, the most popular technique is to add Laplace noise with mean 0 and scale $\Delta f / \epsilon$ where $\Delta f$ is the $L_1$ sensitivity of function $f$. 
Exponential mechanism \citep{McSherry2007} is a general differential privacy mechanism that can be applied to non-numeric outputs.
For categorical data, Dirichlet prior can be used as a noise mechanism to achieve differential privacy \citep{Machanavajjhala2008,McClure2012}.

\subsection{$l$-diversity}

A certain combination of features can identify an individual from an anonymized dataset,
even if personal identifiers, such as driver license number and social security number, are removed from a dataset.
Such threats are commonly prevented by generalizing or suppressing features; for example, ZIP codes with small population are replaced by corresponding county names (generalization), or can be replaced by * (suppression).
Sweeney (\citeyear{Sweeney2002}) proposed a privacy definition for measuring the degree of such feature generalization and suppression, $k$-anonymity.
To adhere the $k$-anonymity principle, each row in a dataset should be indistinguishable with at least $k-1$ other rows.

The definition of $k$-anonymity, however, does not include two important aspects of data privacy: feature diversity and attackers' background knowledge.
Machanavajjhala (\citeyear{Machanavajjhala2007}) illustrated two potential threats to a $k$-anonymized dataset, then proposed a new privacy criterion, $l$-diversity.
The definition of $l$-diversity states that the diversity of sensitive features should be kept within a block of samples. 
There are several ways of achieving $l$-diversity; in this paper, we use Entropy $l$-diversity.
A dataset is Entropy $l$-diverse if 
\begin{align}
- \sum_{x_i} \text{Pr}(x_i \mid \mathbf{x}_{-i}) \log \text{Pr}(x_i \mid \mathbf{x}_{-i}) \ge \log l \label{eq:ldiv}
\end{align}
where $1 \le l $.
This definition originally applies to a dataset with feature generalization or suppression.
For a synthetic sample, Park et al. (\citeyear{Park2013}) suggested an analogous definition of $l$-diversity:
A synthetic dataset is synthetically $l$-diverse if a synthetic sample ${x_i}$ is drawn from a distribution that satisfies $l$-diversity.

\section{Perturbed Gibbs Sampler}\label{sec:pegs}

In this section, we propose the Perturbed Gibbs Sampler (PeGS) for categorical synthetic data.
We first overview the algorithm, then describe its three main components: feature hashing, statistical building blocks, and noise mechanism.
Next, we illustrate how the PeGS algorithm can be efficiently extended to draw a block of random samples.
Finally, we show that multiple imputation can be similarly extended to satisfy differential privacy, which will be used as our baseline model in Section~\ref{sec:empirical}.

\subsection{Algorithm Overview}

\begin{figure}
\begin{center}
\includegraphics[width=0.9\textwidth]{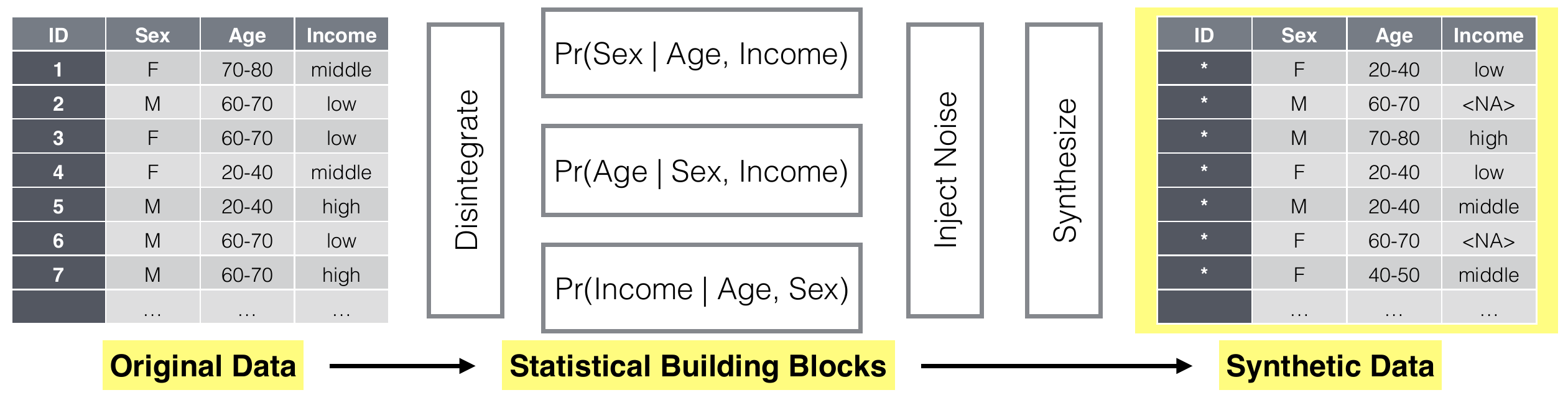}
\end{center}
\caption{PeGS Process Diagram for a three feature dataset: \{(Sex, Age, Race)\}. Three types of conditional distributions are estimated from the original data, then Dirichlet priors are injected to perturb the conditional distributions. Synthetic samples are drawn by iterating over the statistical building blocks.}\label{fig:pegs}
\end{figure}

\begin{figure}
\begin{center}
\includegraphics[width=0.9\textwidth]{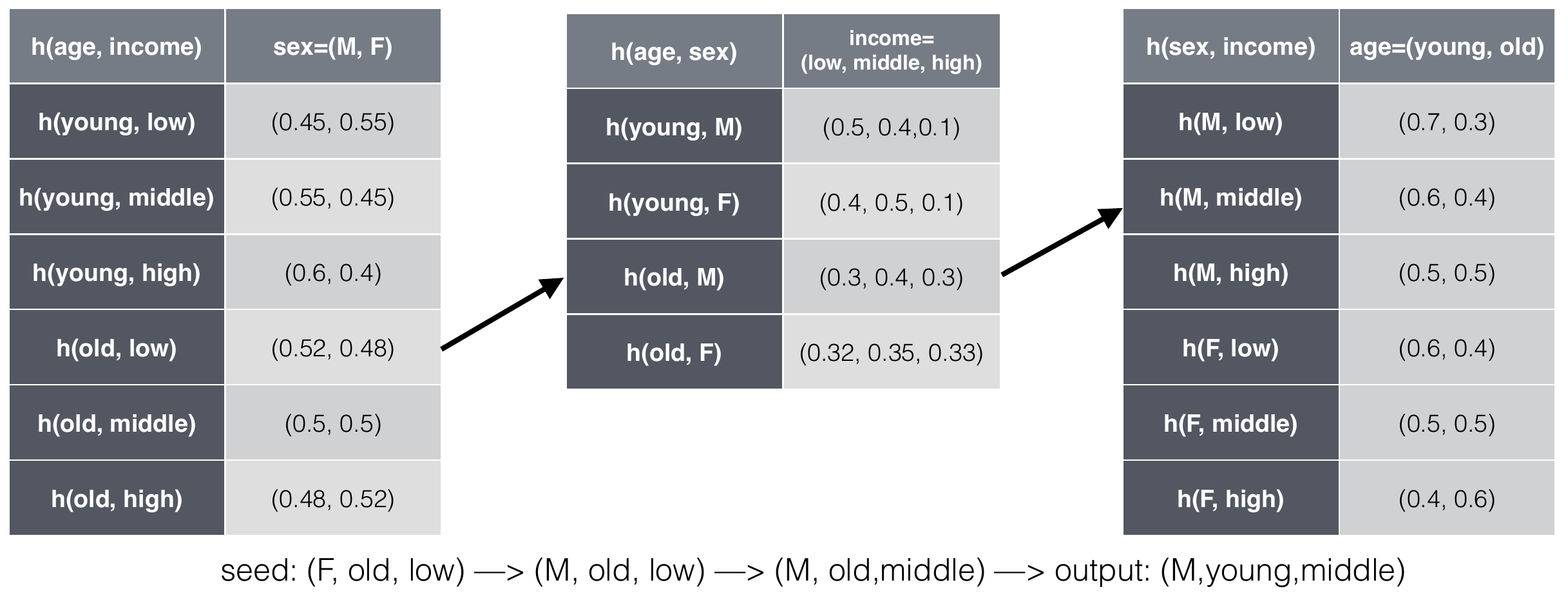}
\end{center}
\caption{Synthesis Steps in PeGS. Three tables represent the statistical building blocks of the example in Figure~\ref{fig:pegs}. In the disintegration step, these three statistical building blocks are stored. In the noise injection step, the probability vectors of the tables are perturbed. In the synthesis step, a new sample is generated by iteratively sampling over the tables. }\label{fig:pegs_sampling}
\end{figure}

Perturbed Gibbs Sampler (PeGS) is a categorical data synthesizer that consists of three main steps:
\begin{enumerate}
\item \textit{Disintegrate}: In this step, the original data $\mathcal{D}$ is disintegrated into statistical building blocks i.e. $\text{Pr}_{\mathcal{D}}(x_i \mid h(\mathbf{x}_{-i}))$ where $h$ is a suitable hash function. These compressed conditional distributions are estimated by counting the corresponding occurrences in the original data.
\item \textit{Inject Noise}: For a specified privacy parameter $\alpha$, the statistical building blocks are modified to satisfy differential privacy or $l$-diversity, $\text{Pr}_{\mathcal{D}}(x_i \mid h(\mathbf{x}_{-i})) \rightarrow \text{Pr}_{\mathcal{D},\alpha}(x_i \mid h(\mathbf{x}_{-i}))$.
\item \textit{Synthesize}: We first pick a random seed from a predefined pool; this can be regarded as a query to our model. The seed sample is transformed to a synthetic sample by iteratively sampling each feature from the statistical building blocks, $x_i \sim \text{Pr}_{\mathcal{D},\alpha}(x_i \mid h(\mathbf{x}_{-i}))$.
\end{enumerate}
Figure~\ref{fig:pegs} visualizes the overall sequential steps of the PeGS algorithm.
Figure~\ref{fig:pegs_sampling} illustrates the synthesis step. 
Three components are essential in the PeGS algorithm: feature hashing, statistical building blocks, and perturbation.
The number of possible conditions is exponential with respect to the number of features,
Therefore, feature hashing is used to compress the number of the possible conditions $\mathbf{x}_{-i}$.
Statistical building blocks are built based on this feature hashing, which are essentially multiple hash-tables describing compressed conditional distributions.
They serve a key role when we try to sample a block of synthetic examples.
Perturbation is required to guarantee the differential privacy. 
Without perturbation, synthetic samples may reveal too much about the original data.                               

\subsection{Feature Hashing}

The hash function $h(\mathbf{x}_{-i})$ in PeGS maps a feature vector to an integer key, where the range of the hash key is much smaller than $2^M$ (exponential in the number of features).
Basically, we want to design a hash function that exhibits good compression while maintaining the statistical properties of data.
The motivation is somewhat similar to feature hashing in machine learning, also known as the hashing trick, which has been often used to compress sparse high-dimensional feature vectors \citep{Weinberger2009}.
For unstructured data such as natural language texts, Locality Sensitive Hashing\citep{Indyk1998} and min-hashing \citep{Gionis1999} can be good candidates for the PeGS hash function.

In this paper, we use a much simpler approach to compress the feature space. 
We order a feature vector $\mathbf{x}_{-i}$ based on the amount of mutual information with $x_i$.
We divide the feature vector into two parts: the first $m \ll M$ number of features and the rest as follows: 
\begin{align*}
\underbrace{x_{o(1)}::\ldots::x_{o(m)}}_{ \text{trivial hash}}::\underbrace{1\text{-bit-hash}({x}_{o(m+1)}::\ldots::{x}_{o(M-1)})}_{\text{1 bit tail}} \longrightarrow \underbrace{1...H}_{\text{Hash key}} 
\end{align*}
where $H \ll 2^M$. 
Let $C_i$ be the number of categories for $x_i$, and $C_{\text{max}} = \max_i C_i$.
The key space of this simple hash function is upper bounded by $2(C_{\text{max}})^m \ll \prod_i C_i$.

The compressed conditional distribution $\text{Pr}(x_i \mid h(\mathbf{x}_{-i}))$, which is basically a occurrence count hash-table for a given hash key, can now be stored in either memory or disk. 
There are several advantages of using this compressed conditional distribution over parametric modeling.
First, the process of building statistical building blocks does not involve complicated statistical procedures such as parameter estimation and model selection.
Second, the resulting statistical building blocks are robust to overfitting. 
Overfitting may occur when there are not enough samples in a table entry. 
Hashing reduces the number of table cells and smoothes out the estimated probability vector. 
Finally, this simple table representation is intuitive, and the process is easily extensible. 
This aspect is critical in our efficient block sampling scheme, which will be illustrated in Section~\ref{sec:rs}.

\subsection{Perturbed Conditional Distribution}\label{sec:perturb}

To satisfy the differential privacy, a certain amount of noise should be injected to the compressed conditional distributions.
The form of noise may depend on applications and privacy measures. 
For example, noise can be added to maximize entropy \citep{Polettini2003} or to satisfy $l$-diversity \citep{Park2013}.
In this paper, we use the Dirichlet prior perturbation to smooth out raw count based estimators to satisfy differential privacy and $l$-diversity.
Specifically, $\alpha$ virtual samples are added to each category of the variable $x_i$, when the conditional distribution $\text{Pr}_\alpha (x_i \mid h(\mathbf{x}_{-i}))$ is estimated.
The amount $\alpha$ is a privacy parameter that controls the degrees of differential privacy and $l$-diversity.
To be more precise, our differentially private perturbation requires a single value of $\alpha$, while our $l$-diverse perturbation needs different $\alpha$ values for each hashed condition $h(\mathbf{x}_{-i})$ i.e. one needs to index $\alpha$ as $\alpha_{h(\mathbf{x}_{-i})}$.
This reflects the fact that differential privacy is a property of the random function, while $l$-diversity depends on dataset properties, an issue that will be touched upon later on.
For analytical simplicity, we assume $\alpha$ virtual samples, $\alpha_{h(\mathbf{x}_{-i})}$ virtual samples for $l$-diversity, are uniformly added to all the categories of the variable $x_i$ (see Equation~\ref{eq:alpha}).
In practice, different amounts of virtual samples can be added to different categories of the variable $x_i$; for example, $\alpha$ can be proportional to the corresponding marginal distribution i.e. $\alpha_j \propto \text{Pr}(x_i=j)$. 

We first derive the probability of sampling $\mathbf{x}$ from the PeGS algorithm.
From a random seed sample $\mathbf{s}$ (or a query), the probability of synthesizing $\mathbf{x}$ is factorized as follows:
\begin{align*}
\text{Pr}_{\mathcal{D}_1, \alpha}(\mathbf{x} \mid \mathbf{s} ) = 
\prod_{i=1}^{M} \text{Pr}_{\mathcal{D}_1, \alpha} (x_i \mid h({x}_{1:(i-1)}, s_{(i+1):M}) )
\end{align*} 
where $x_{1:0}$ and $s_{(M+1):M}$ are just null values. 
For another dataset $\mathcal{D}_2$ that differs by at most one element, the probability of sampling $\mathbf{x}$ can be similarly derived.

For differential privacy (see Equation~\ref{eq:diffpriv}), the ratio between two quantities should satisfy the following relation:
\begin{align*}
\frac{\text{Pr}_{\mathcal{D}_1, \alpha}(\mathbf{x} \mid \mathbf{s})}{\text{Pr}_{\mathcal{D}_2, \alpha}(\mathbf{x} \mid \mathbf{s})}
= \frac{\prod_{i=1}^{M} \text{Pr}_{\mathcal{D}_1, \alpha} (x_i \mid h({x}_{1:(i-1)}, s_{(i+1):M}) )}{\prod_{i=1}^{M} \text{Pr}_{\mathcal{D}_2, \alpha} (x_i \mid h({x}_{1:(i-1)}, s_{(i+1):M}) )}
\le \exp(\epsilon)
\end{align*}
Let us focus on the $i$th component as follows:
\begin{align}
&\text{Pr}_{\mathcal{D}_1, \alpha}(x_i = j \mid h(\mathbf{x}_{-i}) ) = \frac{n_{ij} + \alpha}{N_{h(\mathbf{x}_{-i})} + C_i \alpha} \label{eq:alpha}\\
&N_{h(\mathbf{x}_{-i})} = \sum_{\mathbf{x}_{-i}^\prime} \mathbbm{1}(h(\mathbf{x}_{-i}^\prime)= h(\mathbf{x}_{-i}))
\end{align} 
where $N_{h(\mathbf{x}_{-i})}$ is the total number of rows that have the same hash key as $h(\mathbf{x}_{-i})$ and $n_{ij}$ is the count of the $j$th category i.e. $x_i = j$ within the $N_{h(\mathbf{x}_{-i})}$ samples. 
In other words, the probability of sampling the $j$th category is proportional to the number of the original samples that have the $j$th category.
The privacy parameter $\alpha$ acts as a uniform Dirichlet prior on this raw multinomial count estimate. 

The value of $\alpha$ depends on the privacy criterion. We study two cases: differential privacy and $l$-diversity.

\textbf{A. Differential Privacy. }
The two datasets for defining differential privacy $\mathcal{D}_1$ and $\mathcal{D}_2$ have at most one different row.
Let us assume that $\mathcal{D}_1$ has one more row than $\mathcal{D}_2$ i.e. $\mathcal{D}_1 = \mathcal{D}_2 \cup \mathbf{x}^d$. 
Except for the entry with hash key $h(\mathbf{x}^d_{-i})$, the other entries of the two hash tables from $\mathcal{D}_1$ and $\mathcal{D}_2$ are identical; only one entry of the hash table is different. 
For the different entries of the hash tables, there are two possibilities:
\begin{align*}
\text{if}\quad x_i^d \neq j, \quad&\quad\text{Pr}_{\mathcal{D}_1, \alpha}(x_i^d = j \mid h(\mathbf{x}_{-i}^d) )= \frac{n_{ij} + \alpha}{N_{h(\mathbf{x}_{-i})} + 1+ C_i \alpha}\\
\text{if}\quad x_i^d = j, \quad&\quad \text{Pr}_{\mathcal{D}_1, \alpha}(x_i^d = j \mid h(\mathbf{x}_{-i}^d) )= \frac{n_{ij} + 1 + \alpha}{N_{h(\mathbf{x}_{-i})} + 1+ C_i \alpha}
\end{align*}
Given $\alpha > 0$, we obtain the upper-bound for the $i$th component as follows:
\begin{align*}
\max_{\mathcal{D}_1, \mathcal{D}_2} \frac{\text{Pr}_{ \mathcal{D}_1, \alpha}(x_i =j \mid h(\mathbf{x}_{-i}) )}{\text{Pr}_{\mathcal{D}_2, \alpha}(x_i =j\mid h(\mathbf{x}_{-i}))}
&\le \max_{\mathcal{D}_1, \mathcal{D}_2} \frac{(n_{ij}+1 + \alpha)/(N_{h(\mathbf{x}_{-i})}+1+C_i \alpha)}{(n_{ij} + \alpha)/(N_{h(\mathbf{x}_{-i})}+C_i \alpha)}  
\le 1+\frac{1}{\alpha}
\end{align*}
where the first inequality is because the two datasets only differ by at most one element. 
The second inequality comes from the fact that  $\frac{N_{h(\mathbf{x}_{-i})}+C_i \alpha}{N_{h(\mathbf{x}_{-i})}+1+C_i \alpha} < 1$ and that the equation is maximized when $n_{ij}=0$.
Therefore, we obtain the relation between $\alpha$ and $\epsilon$ as follows:
\begin{align*}
M \log(1 + \frac{1}{\alpha}) \le \epsilon
\end{align*}
Rearranging the terms, we have:
\begin{align}
{\alpha} \le \frac{1}{\exp(\epsilon /M ) - 1} \label{eq:bound}
\end{align}
Note that for univariate binary synthetic data, \citep{McClure2012} showed the relationship between $\alpha$ and $\epsilon$ as $\alpha = \frac{1}{\exp(\epsilon)-1}$.
Equation~(\ref{eq:bound}) says that a higher level of privacy (low $\epsilon$) needs a high value of $\alpha$.
Intuitively, high values of $\alpha$ mean stronger priors, thus the synthetic data are more strongly masked by the priors (or virtual samples).

\textbf{B. $l$-Diversity.}
For $l$-diversity (See Equation~\ref{eq:ldiv}), perturbed conditional distributions need to satisfy the synthetic $l$-diversity criterion:
\begin{align*}
H_\alpha(x_i \mid \mathbf{x}_{-i}) = -\sum_j \text{Pr}_{\mathcal{D},\alpha} \log \text{Pr}_{\mathcal{D},\alpha} \ge \log l
\end{align*}
where $H_\alpha(x_i \mid \mathbf{x}_{-i})$ is the Shannon entropy of the perturbed distribution, $\text{Pr}_{\mathcal{D},\alpha}$.
The entropy $H_\alpha$ is a monotonically increasing function with respect to $\alpha$.
To satisfy the synthetic $l$-diversity criterion with minimal perturbation, we set $\alpha$ as follows:
\begin{align*}
\alpha =
\begin{cases}
    \alpha^* \quad \text{s.t.} \quad  -\sum_j \text{Pr}_{\mathcal{D},\alpha} \log \text{Pr}_{\mathcal{D},\alpha} = \log l, & \text{if}\quad H_\alpha < \log l\\
    0,              & \text{otherwise}
\end{cases}
\end{align*}
where $\alpha$ is set to zero when $H_\alpha$ already satisfies the $l$-diversity criterion.
Unlike the single $\alpha$ for differential privacy, the $\alpha$ values for $l$-diversity vary depending on conditional distributions.
This is because $l$-diversity applies to a dataset, whereas differential privacy applies to a function. 
$l$-diversity is data-aware, but may not provide rigorous guarantees for privacy.
This is also noted in~\citep{Clifton2013} who observed that syntactic methods such as $k$-anonymity and $l$-diversity are designed for privacy-preserving data publishing, while differential privacy is typically applicable for privacy-preserving data mining. 
Thus these two approaches are not directly competing, and indeed can be used side-by-side. 
Clifton and Tassa (\citeyear{Clifton2013}) also provides a detailed assessment of both the limitations and promise of both types of approaches.

\subsection{Removing Sampling Footprints}\label{sec:rs}

\begin{figure}
\begin{center}
\includegraphics[width=0.9\textwidth]{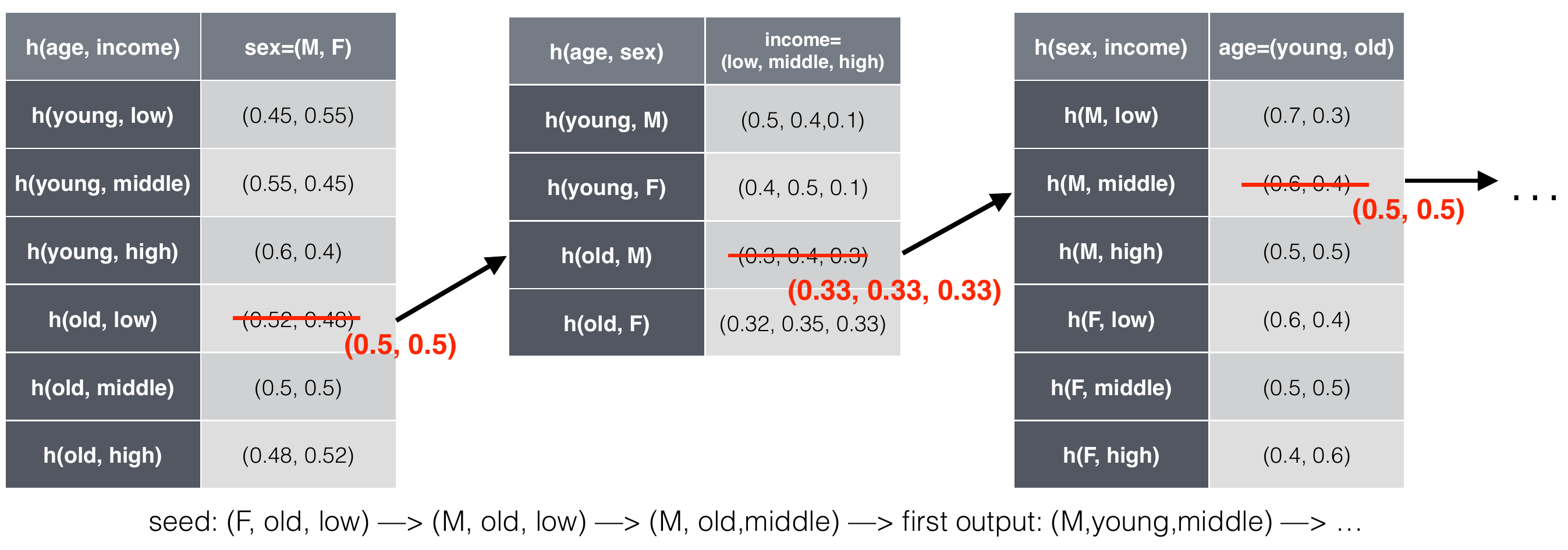}
\end{center}
\caption{Synthesis Steps in PeGS with Reset.  Visited rows in statistical building blocks are reset to the initial state. In this example, the initial states are uniform distributions over categories.}\label{fig:pegs_bsampling}
\end{figure}

This section illustrates an effective block sampling extension of PeGS, and is specific to differential privacy.
PeGS generates one synthetic sample for one seed sample. In other words, one synthetic sample costs $\epsilon$ in the differential privacy regime.
We modify the PeGS algorithm to sample a block of samples from one seed sample, while achieving the same $\epsilon$-differential privacy.
One sampling iteration of PeGS is now repeated many times, but each time, the visited conditional distributions are reset. 
The procedure of Block PeGS with Reset (PeGS.rs) is as follows:
\begin{enumerate}
\item Pick a random seed $\mathbf{s}$ from a predefined pool.
\item For $b$ in $1:B$:
	\begin{enumerate}
	\item Sample $\mathbf{x}^{(b)}$ using PeGS seeded by the previous sample $\mathbf{x}^{(b-1)}$, where $\mathbf{x}^{(0)} = \mathbf{s}$
	\item Reset all visited conditional distributions $\text{Pr}(x_i \mid h(\mathbf{x}_{-i}))$ to uniform distributions
	\end{enumerate}
\end{enumerate}
This algorithm produces a block of synthetic samples $(\mathbf{x}^{(1)}, \ldots, \mathbf{x}^{(B)})$ with the same privacy cost $\epsilon$.
Figure~\ref{fig:pegs_bsampling} illustrates the process of PeGS with Reset.

To analyze the privacy aspect of this modified PeGS algorithm, we first need to calculate the probability of synthesizing a block of samples:
\begin{align*}
\text{Pr}_{\mathcal{D}_1,\alpha}^{B} ( \mathbf{x}^{(1)}, \ldots, \mathbf{x}^{(B)} \mid \mathbf{s} ) = \text{Pr}_{\mathcal{D}_1,\alpha}^{(1)} ( \mathbf{x}^{(1)} \mid \mathbf{s} ) \prod_{b=2}^B \text{Pr}_{\mathcal{D}_1,\alpha}^{(b)} ( \mathbf{x}^{(b)} \mid \mathbf{x}^{(b-1)} )
\end{align*}
where $ \text{Pr}_{\mathcal{D}_1,\alpha}^{(b)} ( \mathbf{x}^{(b)} \mid \mathbf{x}^{(b-1)} )$ is the transition probability from $\mathbf{x}^{(b-1)}$ to $\mathbf{x}^{(b)}$.
Note that $\text{Pr}^{(b)}$ and $\text{Pr}^{(b+1)}$ are different conditional distributions, as $M$ components of $\text{Pr}^{(b)}$ are reset to the initial states.
The ratio between two probabilities is written as follows:
\begin{align*}
\frac{\text{Pr}_{\mathcal{D}_1,\alpha}^{(1)} ( \mathbf{x}^{(1)} \mid \mathbf{s} ) \prod_{b=2}^B \text{Pr}_{\mathcal{D}_1,\alpha}^{(b)} ( \mathbf{x}^{(b)} \mid \mathbf{x}^{(b-1)} )}{\text{Pr}_{\mathcal{D}_2,\alpha}^{(1)} ( \mathbf{x}^{(1)} \mid \mathbf{s} ) \prod_{b=2}^B \text{Pr}_{\mathcal{D}_2,\alpha}^{(b)} ( \mathbf{x}^{(b)} \mid \mathbf{x}^{(b-1)} )} \le \exp(\epsilon)
\end{align*}
Recall that the statistical building blocks from both datasets differ at most $M$ components, as the two datasets differ at most one element.
We provide a sketch of the proof that this algorithm satisfies $\epsilon$-differential privacy as follows:
\begin{enumerate}
\item To generate the same block of samples, the sequences of statistical building blocks need to be the same as well. 
In other words, as the two samples, $\mathbf{x}^{(b)} \mid \mathcal{D}_1$ and $\mathbf{x}^{(b)} \mid \mathcal{D}_2$, are the same, $\mathbf{x}^{(b)}_{-i} \mid \mathcal{D}_1$ and $\mathbf{x}^{(b)}_{-i} \mid \mathcal{D}_2$ will also be the same. Thus, they use the building blocks from the same location for sampling $x_i$ at the $b$th iteration, $\text{Pr}^{(b)}_{\mathcal{D}_1, \alpha}(x_i \mid \mathbf{x}_{-i})$ and $\text{Pr}^{(b)}_{\mathcal{D}_2, \alpha}(x_i \mid \mathbf{x}_{-i})$.
\item There are at most $M$ different components between $\text{Pr}^{(1)}_{\mathcal{D}_1,\alpha}$ and $\text{Pr}^{(1)}_{\mathcal{D}_2,\alpha}$, and let $\mathcal{M}$ be the set of different components. This is because $\mathcal{D}_1$ and $\mathcal{D}_2$ differ by at most one row.
\item If $\text{Pr}^{(1)}_{\mathcal{D}_1,\alpha}$ touched $(M-d)$ components in $\mathcal{M}$, then $\frac{(M-d)}{M}\epsilon$ privacy cost is spent in the process (see Section~\ref{sec:perturb}). 
\item If $\text{Pr}^{(1)}_{\mathcal{D}_1,\alpha}$ touched $(M-d)$ components in $\mathcal{M}$, then the rest of the sequences can differ at most $d$ components. This is because those $(M-d)$ components are reset to uniform distributions, and they became indistinguishable i.e. the visited components from $\mathcal{D}_1$ and $\mathcal{D}_2$ became the same uniform distribution. Every visit of an element in $\mathcal{M}$ decreases the number of different elements. 
\item Therefore, the whole sequence can differ at most $M$ components (upper-bound), thus the proposed block sampling algorithm satisfies the same $\epsilon$-differential privacy for generating a block of $B$ samples.
\end{enumerate}
As we have more samples for the same cost, the privacy cost per sample can be written as:
\begin{align*}
{\alpha} \le \frac{1}{\exp(\epsilon^\prime B /M ) - 1}
\end{align*}
where $\epsilon/B = \epsilon^\prime$.
As can be seen, the privacy cost is smaller by a factor of $B$.
However, the block size $B$ cannot be arbitrarily large. 
As every visited statistical building block is reset, the synthetic samples tend to be more noisy as we increase the size of the block.
This property will be illustrated using a real dataset in Section~\ref{sec:empirical}.

\subsection{Perturbed Multiple Imputation}

The Dirichlet perturbation similarly can be applied to multiple imputation.
Perturbed Multiple Imputation is a naive extension of multiple imputation that satisfies $\epsilon$-differential privacy.
A multiple imputation with generalized linear models can be written as follows:
\begin{align*}
\text{Pr}_{\hat{\mathbf{w}}(\mathbf{x})}(\mathbf{x} ) = \prod_{i=1}^M  g_{x_i} (\hat{\mathbf{w}}_i(\mathcal{D}_1)^\top \mathbf{x}_{-i}))
\end{align*}
where $g_{x_i} (\hat{\mathbf{w}}_i(\mathcal{D}_1)^\top \mathbf{x}_{-i}))$ is the estimated response probability of $x_i$ using a generalized linear model.
We assume that the response is a \textit{normalized} probability measure, thus $g_{x_i} \in [0, 1]$.
We propose perturbed multiple imputation as follows:
\begin{align*}
\text{Pr}_{\hat{\mathbf{w}}(\mathbf{x}), \alpha}(\mathbf{x} ) = \prod_{i=1}^M  g_{x_i}^\alpha (\hat{\mathbf{w}}_i(\mathcal{D}_1)^\top \mathbf{x}_{-i}))
\end{align*}
Perturbed multiple imputation satisfies $\epsilon$-differential privacy, if the output is perturbed as
\begin{align*}
g_{x_i}^\alpha(\hat{\mathbf{w}}_i(\mathcal{D}_1)^\top \mathbf{x}_{-i}) = \frac{g_{x_i}(\hat{\mathbf{w}}_i(\mathcal{D}_1)^\top \mathbf{x}_{-i}) + \alpha}{\sum_{x_i} g_{x_i \in X_i}(\hat{\mathbf{w}}_i(\mathcal{D}_1)^\top \mathbf{x}_{-i}) + C_i \alpha} = \frac{g_{x_i}(\hat{\mathbf{w}}_i(\mathcal{D}_1)^\top \mathbf{x}_{-i}) + \alpha}{1 + C_i \alpha} 
\end{align*}
where $\alpha = 1/(\exp(\epsilon/M)-1)$. 
The proof is analogous to the proof for the PeGS algorithm.
With $\alpha=0$, this algorithm is the same as a multiple imputation with generalized linear model.

\section{Empirical Study}\label{sec:empirical}

In this section, we evaluate the PeGS algorithm using a real dataset from two perspectives: utility and risk of the PeGS-synthesized data.
The utility is measured by comparing marginal, conditional distributions and regression coefficients with those from the original data.
The risk is first measured by the differential privacy parameter $\epsilon$.
As the differential privacy parameter can be too conservative for a real dataset, we also measure population uniqueness and indirect probabilistic disclosure risks.
The presented experiments are mainly for the differentially private perturbation, and the experiment with the $l$-diversity perturbation can be found in \citep{Park2013}.

\subsection{Dataset Overview}

We use public Patient Discharge Data from California Office of Statewide Health Planning and Development\footnote{\url{http://www.oshpd.ca.gov/HID/Data_Request_Center/Manuals_Guides.html}}.
This dataset contains inpatient, emergency care, and ambulatory surgery data collected from licensed California hospitals.
Each row of the data represents either one discharge event of a patient or one outpatient encounter. 
The data are already processed with several disclosure limitation techniques. 
Feature generalization and masking rules are applied to the data based on population uniqueness.

For our experiment, we use 2011 Los Angeles data.
Although there are almost 40 variables in the provided data, we use 13 important variables.
The selected variables are listed in Table~\ref{tab:data}.
For the numeric variables such as age and charge, we transformed the variables into categorical variables by grouping.
We subset the data to focus on populous zip code areas. 
This is to prevent any possible privacy infringement from our experiment.
We use this preprocessed dataset to be our ground-truth original data.
As can be seen, the possible combinations of the categories are approximately 2 trillion:  $2 \times 10^{12} \approx 6 \times 18 \times 3 \times 4 \times 7 \times 16 \times 16 \times 13 \times 9 \times 25 \times 25 \times 3 \times 2 $.
A table of this size cannot be stored in a personal computer.

Diagnostic and procedural codes are not included in this experiment.
In the original data, diagnoses and procedures are coded following the rules of International Classification of Diseases (ICD-9).
Both codes can specify very fine levels of diagnoses and procedures; for example, the ICD-9 codes include information about a underlying disease and a manifestation in a particular organ.
These diagnostic and procedural codes can be grouped into a smaller number of categories.
Major Diagnostic Categories (MDC) and Medicare Severity Diagnosis-Related Group (MSDRG) are two examples of coarser diagnostic codes.
In this example, we only include higher level abstractions of the detailed features.
To keep the semantics of the data, we recommend a two step procedure: first generating a higher level feature, then synthesizing detailed features based on the higher level feature.

Three numeric variables, \texttt{age}, length-of-stay (\texttt{los}), and \texttt{charge}, are grouped and transformed into categorical features.
The age variable is equipartitioned to have 5 years gap between consecutive categories.
The \texttt{los} and \texttt{charge} variables are grouped based on their marginal distributions.
For example, almost half of the population stayed less than 10 days in a hospital.
Thus, the \texttt{los} variable is grouped to have 1 day gap before 10 days threshold, and 20 days gap after 10 days.
The \texttt{charge} variable exhibited a similar marginal distribution; almost a half of the population pay less than 20K dollars, and we binned this variable to have almost equal sizes of population.
The grouping rules are illustrated in Table~\ref{tab:data}.
In Section~\ref{sec:conclusion}, we will discuss the limitations and extension of treating numeric variables in the PeGS framework.

\begin{table}\caption{California discharge data. Los Angeles. }\label{tab:data}
\begin{center}
\begin{scriptsize}
\begin{tabular}{| c | c | c | }\hline
Variable Name   & Description & Category Values \\\hline
typ & Type of care & Acute Care, Skilled Nursing, Psychiatric, etc. (6 levels)\\
age.yrs & Age of the patient (5 years bin)  & 0, 5, 10, 15, ..., 80, NA (18 levels) \\
sex & Gender of the patient & Male, Female, NA (3 levels)\\
ethncty & Ethnicity of the patient & Hispanic, Non-Hispanic, Unknown, NA (4 levels) \\
race & Race of the patient & White, Black, Native American, Asian, etc. (7 levels) \\
patzip & Patient ZIP code (in LA) & 900xx, 902xx, ... , 935xx (16 levels) \\
los & Length of stay (in days) & 0, 1, 2, ... , 9, 10-30, 30-50, 50-70, 90+, NA (16 levels) \\
disp & The consequent arrangement & Routine, Acute Care, Other Care, etc. (13 levels) \\
pay & Payment category& Meicare, Medi-Cal, Private, etc. (9 levels) \\
charge & Total hospital charge during the stay & 0, 2K, 6K, 8K, 10K, 15K, 20K, ..., 100K+ (25 levels) \\
MDC & Major diagnostic category & Nervous sys., Eye, ENMT, etc. (25 levels) \\
sev & Severity code  & 0, 1, 2 (3 levels) \\
cat & Category code & Medical, Surgical (2 levels) \\\hline
\end{tabular}
\end{scriptsize}
\end{center}
\end{table}

\subsection{Sampling Demonstration}

PeGS transforms each feature one by one conditioned on the rest of the features.
This approach differs from a multiple imputation strategy in two aspects.
First, PeGS estimates compressed conditional distributions rather than parameterized approximations e.g., generalized linear models.
Second, the compressed conditional distributions can be further perturbed by calibrating the privacy parameter, which makes synthetic data $\epsilon$-differentially private.
Table~\ref{tab:pegs_it_demo} shows how PeGS transforms a random seed into a private synthetic sample. 
The first row of the table is a random seed, and each consecutive row shows the corresponding sampling step.
Note that some features change their values, whereas other features maintain the original values. 
The final sample is shown in the last row. 
As can be seen, the final transformed sample is different from the seed; for example, it has a different age, zip code, and disposition code.

\begin{table}\caption{Detailed Sampling Steps in the PeGS synthesis step. Four variables, sex, race, payment category (pay), category code (cat), are not changed in the final transformed example.}\label{tab:pegs_it_demo}
\begin{center}
\begin{scriptsize}
\begin{tabular}{| c | c | c | c | c | c | c | c | c | c | c | c | c | c | }\hline
sequence   & typ & age.yrs & sex & ethncty & race & patzip & los & disp & pay & charge & MDC & sev & cat \\\hline
seed      & 4 & 55 &  2  & 1 & 1 & 917xx & 8 &1 & 3 & 40K  & 25  &  1 &  M\\
$X_1 \mid X_{-1}$ & \textit{5} &  55 &  2 &      1  &  1 &   917xx &   8  &  1 &      3 & 40K  & 25      &  1  &      M \\
$X_2 \mid X_{-2}$  & 5 &  \textit{75} &  2  &     1  &  1 &   917xx  & 8  &  1    &   3 & 40K & 25     &   1    &    M \\
$X_3 \mid X_{-3}$        & 5 &  75 &  \textit{2}    &   1  &  1 &   917xx  & 8  &  1   &    3 & 40K & 25     &   1   &     M \\
$X_4 \mid X_{-4}$  & 5 &  75 &  2  &     \textit{2}  &  1 &   917xx  & 8  &  1    &   3 & 40K & 25     &   1    &    M \\
$X_5 \mid X_{-5}$       & 5 &  75 &  2   &    2  &  \textit{1} &   917xx  & 8  &  1    &   3 & 40K & 25    &    1    &    M \\
$X_6 \mid X_{-6}$    & 5 &  75 &  2     &  2  &  1 &   \textit{913xx}  & 8  &  1    &   3 & 40K & 25    &    1    &    M \\
$X_7 \mid X_{-7}$         & 5 &  75 &  2    &   2  &  1 &   913xx  & \textit{9}  &  1   &    3 & 40K & 25    &    1   &     M \\ 
$X_8 \mid X_{-8}$       & 5 &  75 &  2    &   2   & 1 &   913xx   &9  &  \textit{5}   &    3 & 40K & 25    &    1   &     M \\
$X_9 \mid X_{-9}$  & 5 &  75 &  2    &   2   & 1 &   913xx  & 9  &  5    &   \textit{3} & 40K & 25    &    1    &    M  \\
$X_{10} \mid X_{-10}$  & 5 &  75 &  2    &   2  &  1  &  913xx  & 9  &  5    &   3 & \textit{65K} & 25    &    1    &    M \\
$X_{11} \mid X_{-11}$     & 5 &  75 &  2    &   2  &  1 &   913xx  & 9  &  5   &    3 & 65K  & \textit{7}    &    1    &    M \\ 
$X_{12} \mid X_{-12}$ & 5 &  75 &  2   &    2  &  1 &   913xx  & 9  &  5   &    3 & 65K &  7   &     \textit{0}    &    M \\
$X_{13} \mid X_{-13}$ & 5 &  75 &  2   &   2  &  1  &  913xx  & 9 &   5    &   3 & 65K  & 7  &      0      &  \textit{M} \\\hline
\end{tabular}
\end{scriptsize}
\end{center}
\end{table}

\begin{figure}[h]
\begin{center}
\begin{subfigure}[b]{0.45\textwidth}
	\includegraphics[width=1\textwidth]{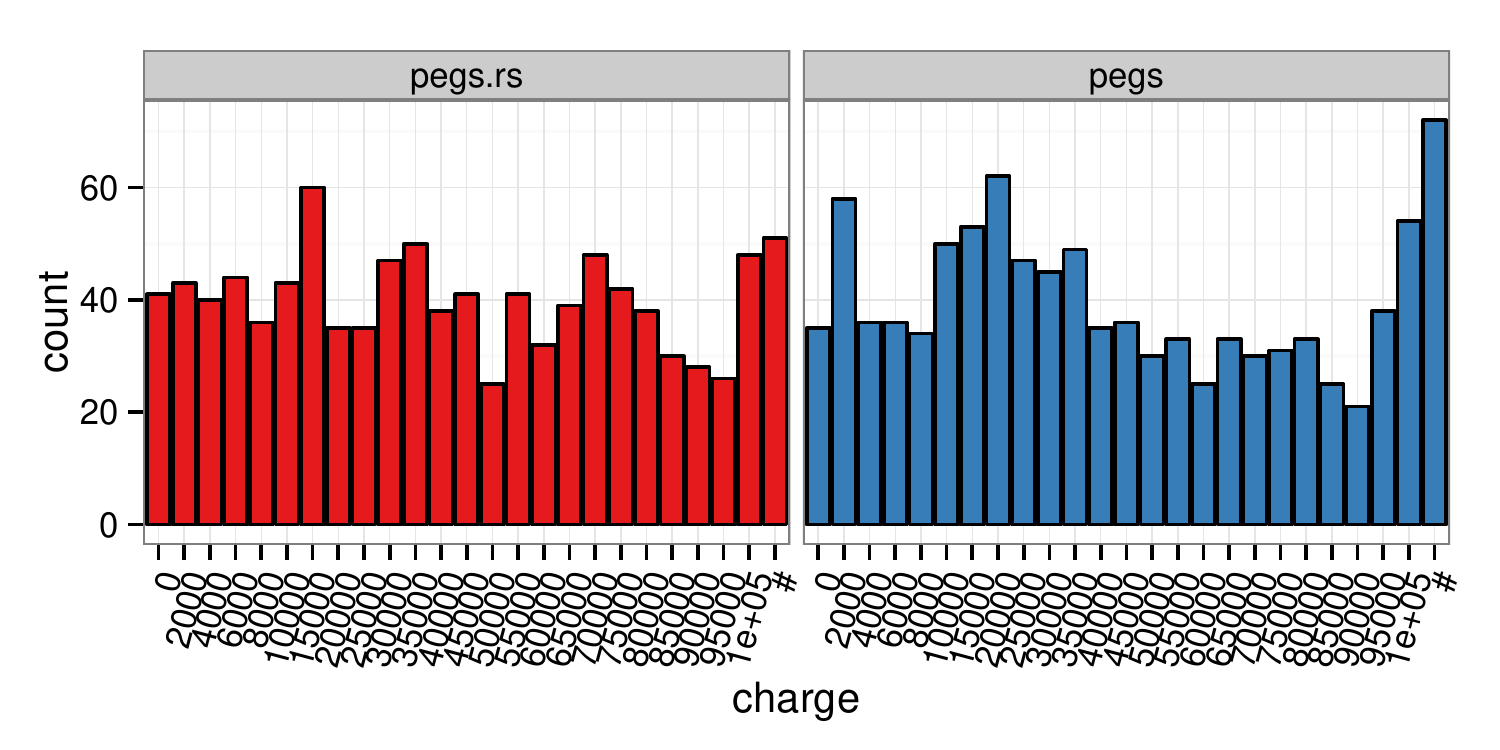}
	\caption{Histogram on the charge variable}
\end{subfigure}
\begin{subfigure}[b]{0.45\textwidth}
	\includegraphics[width=1\textwidth]{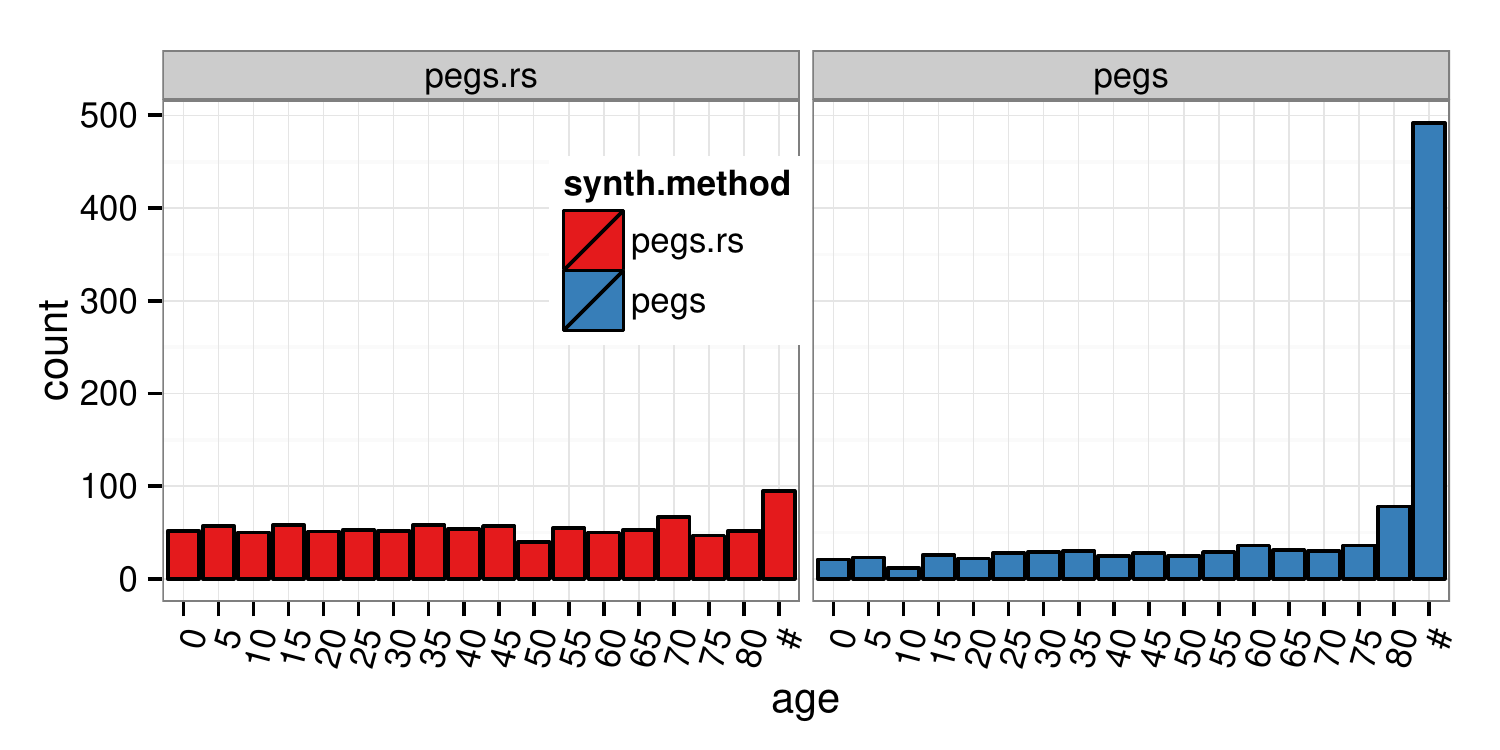}
	\caption{Histogram on the age variable}
\end{subfigure}
\end{center}
\caption{Histograms for comparing PeGS and PeGS.rs (PeGS with Reset). }\label{fig:pegs_rs_hist_compare}
\end{figure}

Unlike multiple imputation, PeGS can be iterated many times.
However, without the reset option, there is no gain for the privacy cost.
The reset option in PeGS.rs removes sampling footsteps, but the synthetic samples after many iterations may not be useful for representing the original data.  
Figure~\ref{fig:pegs_rs_hist_compare} shows histograms from the generated samples. 
As can be seen, the block samples from PeGS.rs are more uniformly distributed than those from PeGS.
The distributions from PeGS are actually closer to the distribution of the original data than those from PeGS.rs.
It is important to note that the PeGS and PeGS.rs in this experiment have different privacy cost; PeGS.rs only used $\epsilon$, while PeGS requires $\epsilon \times \text{Iterations}$.
The goal of this experiment is to show the limitation of PeGS.rs.
Although PeGS.rs provides more number of samples given the same privacy cost, an arbitrarily large size of block may not be useful in practice.

\subsection{Risk ($\epsilon$) vs. Utility}

Reducing disclosure risk and improving data utility are two competing objectives when publishing privacy-safe synthetic data.
As these two goals cannot be satisfied at the same time, a certain trade-off is necessary for preparing public use data.
This trade-off has been traditionally represented using a graphical measure, called R-U confidentiality map \citep{Duncan2001}.
The R-U confidentiality map consists of two axis: typically a risk measure on the x-axis and a utility measure on the y-axis.
Note that risk and utility measures can be domain and application specific. 
In this paper, we first show R-U maps where the risk is measured using differential privacy.
The utility is primarily measured by comparing statistics from the original data and synthetic data.

We use three different algorithms and seven different privacy parameters for each algorithm as follows:
\begin{itemize}
\item PeGS: Perturbed Gibbs Sampler
\item PeGS.rs: Perturbed Gibbs Block Sampler with Reset. Block size = 10.
\item PMI: Perturbed Multiple Imputation (baseline algorithm). With higher values of $\epsilon$, this is the same as a multiple imputation strategy for fully synthetic data. 
In PMI, the conditional distributions are modeled using the elastic-net regularized multinomial logistic regression, specifically \texttt{glmnet} package in \texttt{R} 2.15.3 \citep{Friedman2010}. 
The variable $x_i$ is regressed on the rest of the variables $\mathbf{x}_{-i}$, and the regularization parameter $\lambda$ was tuned based-on cross-validation:
\begin{align*}
\text{Pr}(x_i=j \mid \mathbf{x}_{-i}) \propto \exp(c_{ij} + \boldsymbol{\beta}_{ij}^\top \mathbf{x}_{-i})
\end{align*}
where $c_{ij}$ and $\boldsymbol{\beta}_{ij}$ are estimated from the data.
\end{itemize}
where the privacy parameters are given as $\epsilon \in \{0.1, 0.5, 1, 5, 10, 50, 100\}$ per synthetic sample. We generated 1000 samples for each case.
As a result, we have $21 = 7 \times 3$ synthetic datasets and one original dataset.

\begin{figure}
\begin{center}
\includegraphics[width=0.8\textwidth]{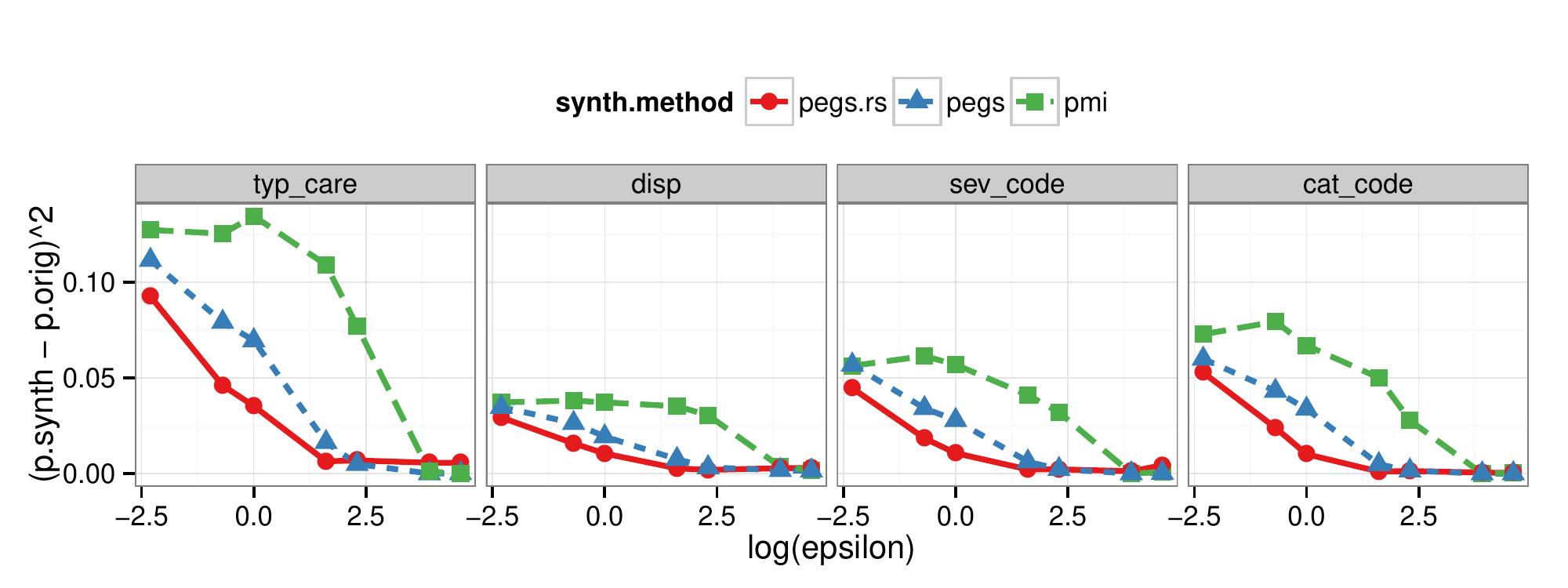}
\end{center}
\caption{R-U maps where the utility is measured as the difference in marginal distributions.}\label{fig:ru_marginals}
\end{figure}

\begin{figure}[h]
\begin{center}
\begin{subfigure}[b]{0.8\textwidth}
	\includegraphics[width=1\textwidth]{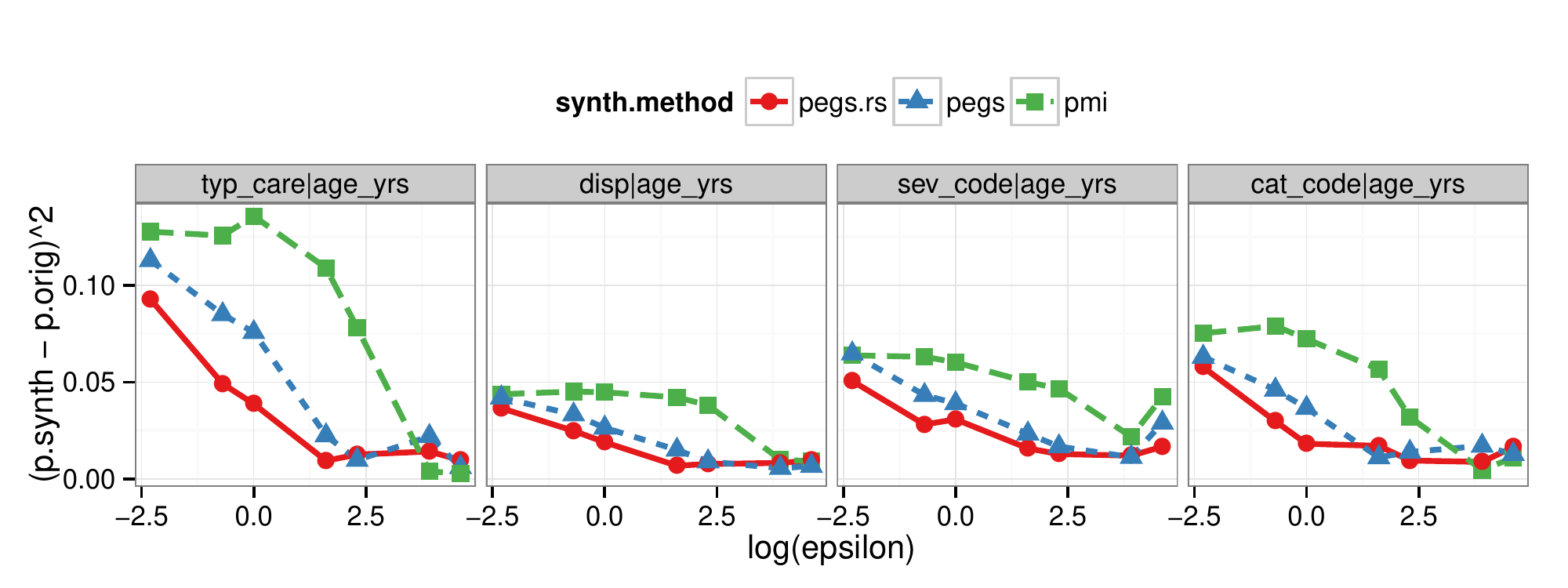}
	\caption{Distributional distances conditioned on the age variable, $X_i \mid \text{age.yrs}$}
\end{subfigure}

\begin{subfigure}[b]{0.8\textwidth}
	\includegraphics[width=1\textwidth]{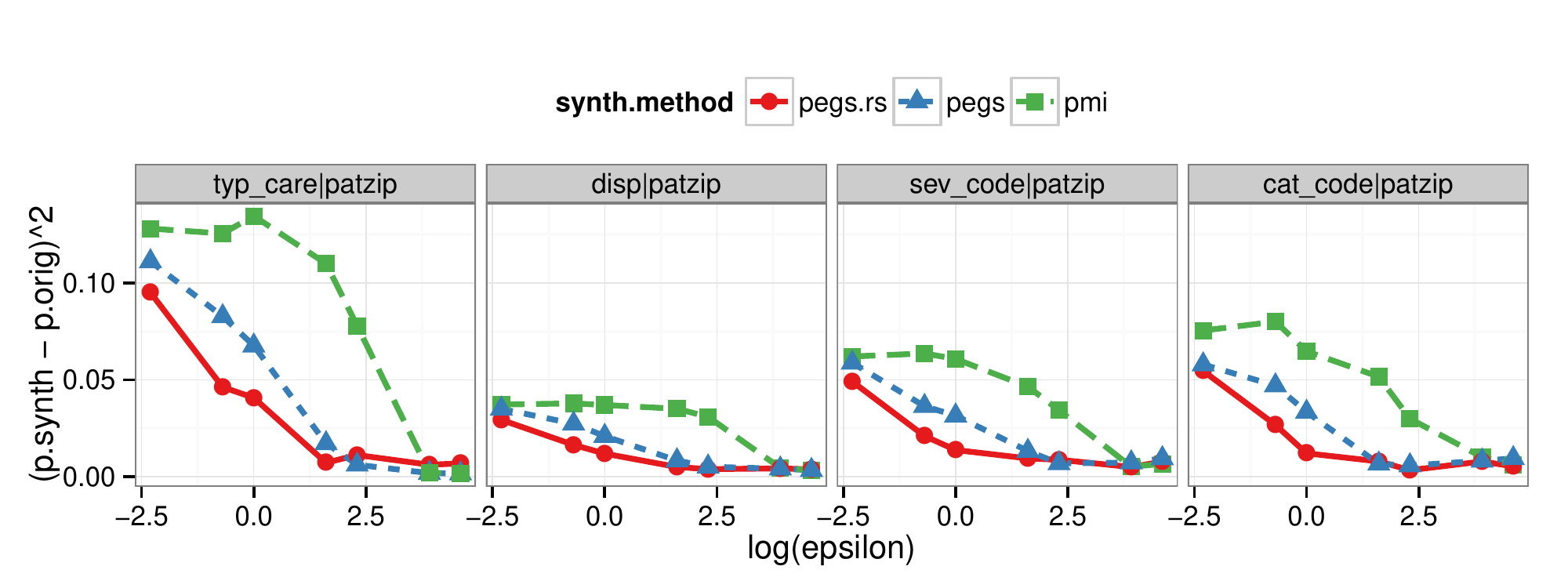}
	\caption{Distributional distances conditioned on the ZIP code variable, $X_i \mid \text{patzip}$}
\end{subfigure}
\end{center}
\caption{R-U maps where the utility is measured as the difference in conditional distributions.}\label{fig:ru_conditionals}
\end{figure}

The utility is first measured using marginal and conditional distributions.
Marginal and conditional distributions are measured from the original and synthetic datasets, then the distance is calculated as follows:
\begin{align*}
\text{Marginal Distance} &= \sum_{x_i \in X_i} (\hat{\text{Pr}}_{\text{synth},\epsilon}(x_i ) - \hat{\text{Pr}}_{\text{orig}}(x_i ))^2\\
\text{Conditional Distance} &= \sum_{x_j \in X_j}\sum_{x_i \in X_i} (\hat{\text{Pr}}_{\text{synth},\epsilon}(x_i \mid x_j) - \hat{\text{Pr}}_{\text{orig}}(x_i \mid x_j ))^2
\end{align*}
where the distance is an inverse surrogate for the utility.
Figure~\ref{fig:ru_marginals} and Figure~\ref{fig:ru_conditionals} show the R-U maps where the utility is measured as the difference in marginal and conditional distributions, respectively.
As can be seen, all synthetic datasets become similar to the original data with higher values of $\epsilon$.
However, for smaller values of $\epsilon$, the synthetic data from PeGS.rs are much more similar to the original than the others. 
The distributional distances of PeGS are slightly smaller than those of PeGS.rs for higher values of $\epsilon$.
Since $\alpha$ values are very small for these privacy parameters, the reset operation of PeGS.rs becomes more noticeable, and it pushes synthetic samples away from the original distributions.

\begin{figure}
\begin{center}
\includegraphics[width=0.7\textwidth]{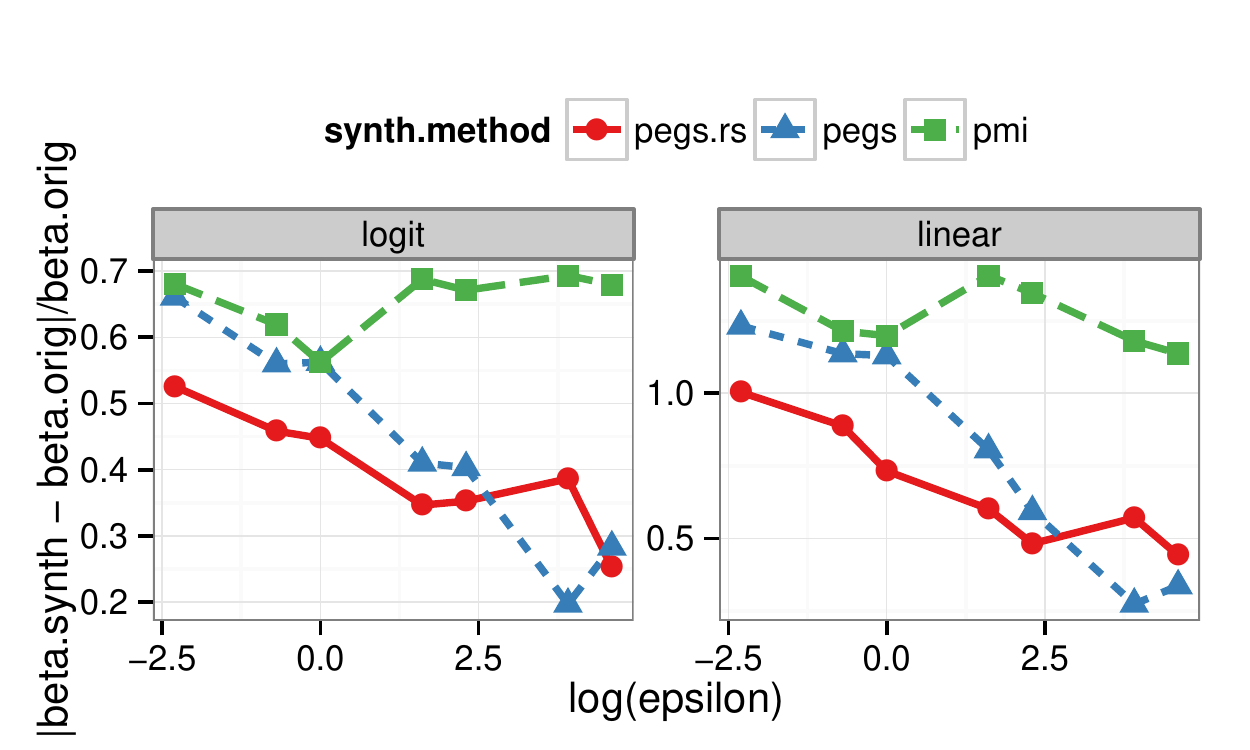}
\end{center}
\caption{R-U maps where the utility is measured as the difference in regression coefficients.}\label{fig:ru_regression}
\end{figure}

\begin{figure}[h]
\begin{center}
\begin{subfigure}[b]{0.45\textwidth}
	\includegraphics[width=1\textwidth]{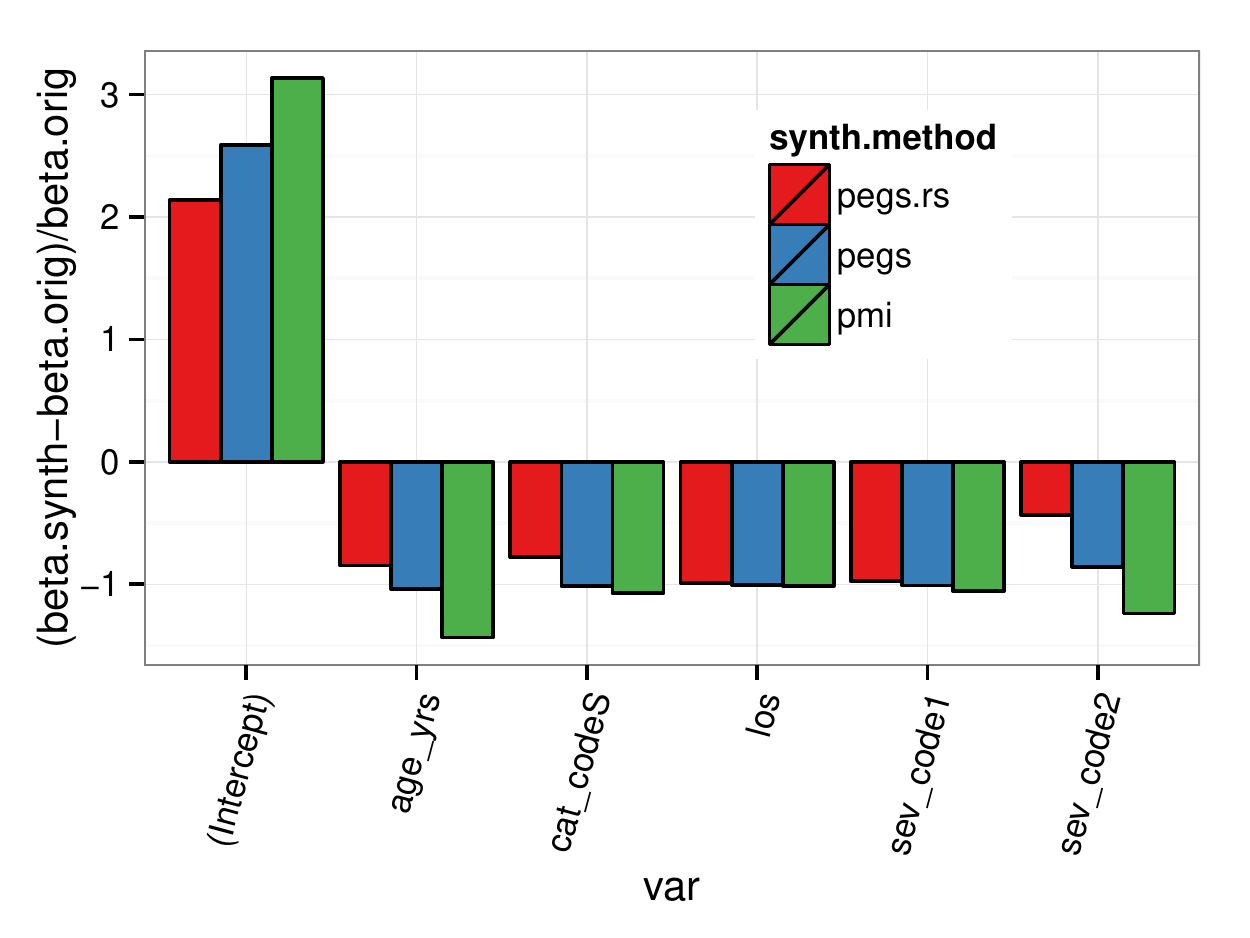}
	\caption{$\epsilon=0.1$}
\end{subfigure}
\begin{subfigure}[b]{0.45\textwidth}
	\includegraphics[width=1\textwidth]{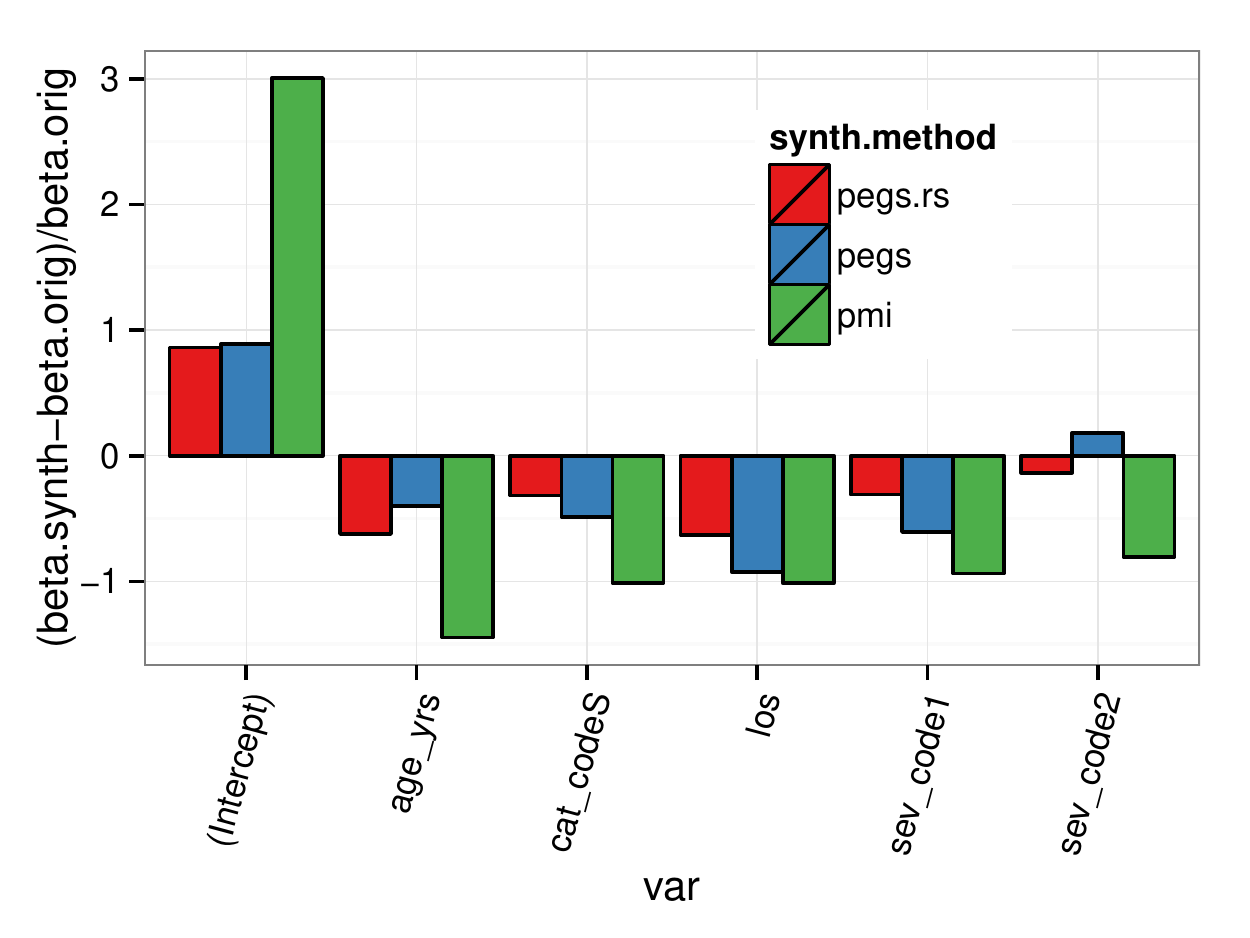}
	\caption{$\epsilon=10$}
\end{subfigure}
\end{center}
\caption{Estimated logistic regression coefficients for (a) $\epsilon=0.1$ and (b) $\epsilon=10$. The coefficients from $\epsilon=10$ (lower level of privacy) are closer to the original coefficients than those from $\epsilon=0.1$ (higher level of privacy). }\label{fig:coef_regression}
\end{figure}

Next, we compare the coefficients from regression models learned on the datasets.
We learned logistic and linear models as follows:
\begin{align*}
I(\text{charge}>25K) &\sim \text{as.numeric}(\text{age.yrs}) + \text{sev} + \text{cat} + \text{as.numeric}(\text{los})\\
\text{as.numeric}(\text{charge}) &\sim \text{as.numeric}(\text{age.yrs}) + \text{sev} + \text{cat} + \text{as.numeric}(\text{los})
\end{align*}
where some of the features are changed to numeric features based on their actual meaning.
The choice of the target variable was arbitrary, as the goal of this illustrative experiment is to show the applicability of synthetic data in predictive modeling tasks.
After learning the coefficients of each model, the distance between the coefficients is measured as follows:
\begin{align*}
\text{Regression Distance} = \sum_{i} | \frac{\beta_{i,\text{synth}} - \beta_{i,\text{orig}}}{\beta_{i,\text{orig}}}|
\end{align*}
Figure~\ref{fig:ru_regression} shows the R-U map from the regression experiment.
As can be seen, the synthetic samples from PeGS.rs provide the most similar coefficients to those from the original data.
Figure~\ref{fig:coef_regression} shows each coefficient deviation from the linear regression example for two different differential privacy levels.
Notice that the intercept coefficients from the synthetic datasets tend to overshoot the actual value, while the other feature coefficients tend to undershoot.
This is because the perturbation decreases all feature correlations including the correlation between the target and independent variables.

\subsection{Estimating Re-identification Risk}

Although differential privacy provides a theoretically sound framework for measuring disclosure risks, the measure is originally designed for functions, not data \citep{Dankar2012}.
For many cases, the measures can be overly conservative or strict for a real dataset.
In the statistical disclosure limitation literature, there have been many attempts to measure disclosure risks for synthetic data.
Franconi and Stander (\citeyear{Franconi2002}) proposed a method to quantify disclosure risks for model-based synthetic data.
Their proposed approach checks whether it is possible to recognize a unit in the released data assuming the original data are given to an intruder.
This provides a somewhat conservative measure, but is still useful to compare the risks from different release mechanisms.
Reiter (\citeyear{Reiter2005b}) later formalized measuring probabilistic disclosure risk scores for partially or fully synthetic data. 
Probabilistic disclosure risks are used to asses the risks of the fully synthetic data using Random Forests in Caiola and Reiter (\citeyear{Caiola2010}).

\begin{figure}
\begin{center}
\begin{subfigure}[b]{0.3\textwidth}
	\includegraphics[width=1\textwidth]{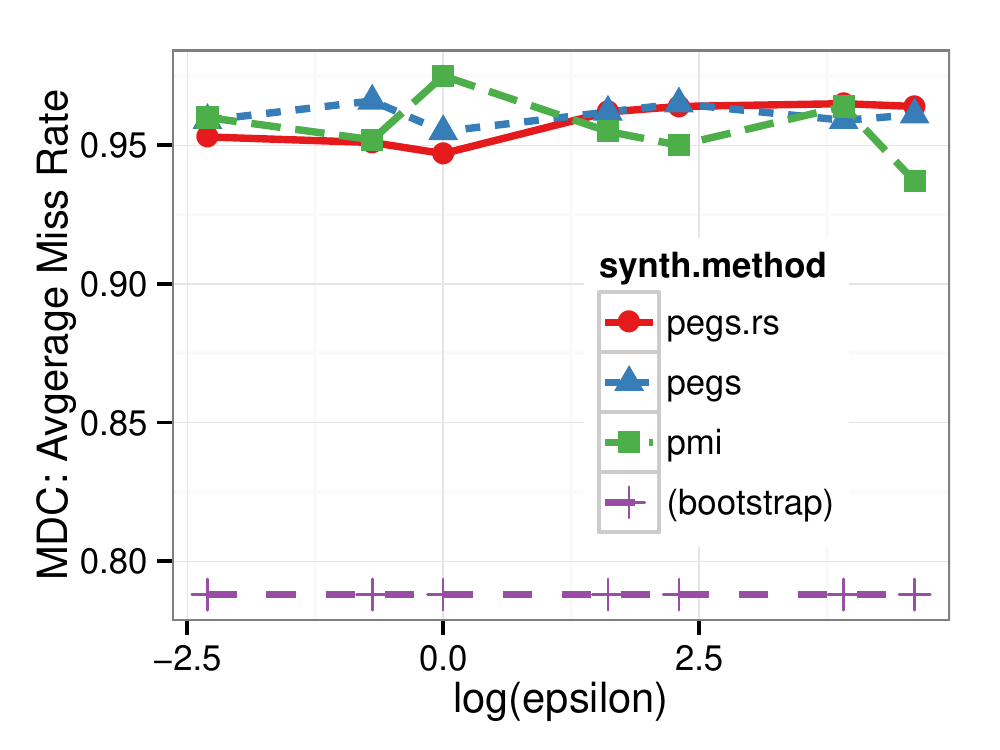}
	\caption{Attack on the MDC variable}
\end{subfigure}
\begin{subfigure}[b]{0.3\textwidth}
	\includegraphics[width=1\textwidth]{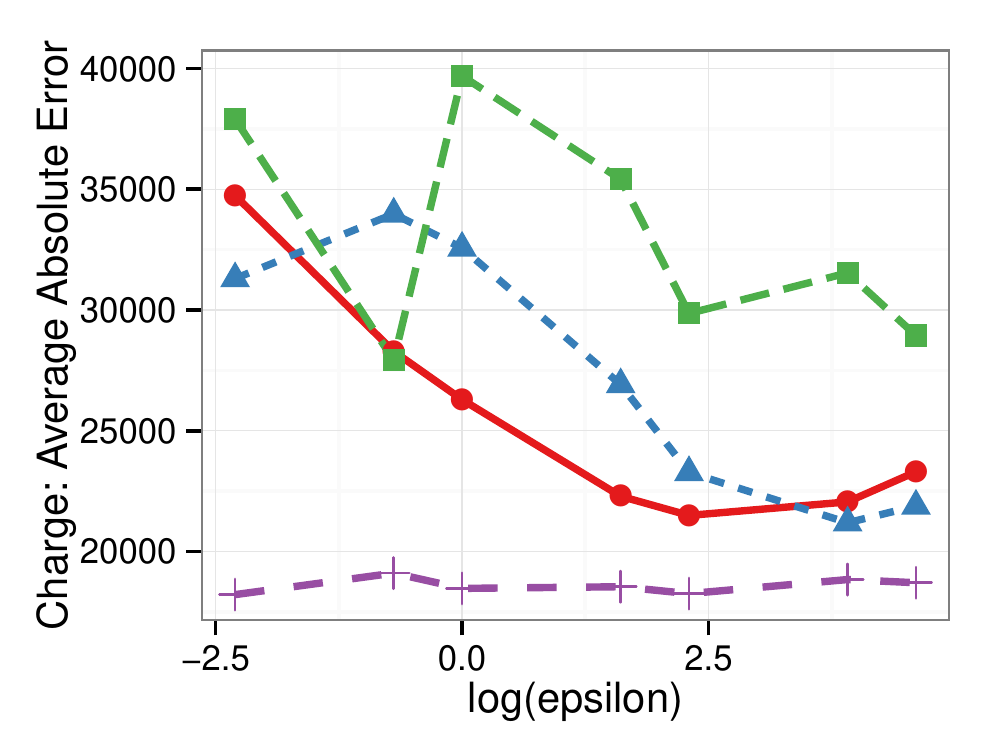}
	\caption{Attack on the charge variable}
\end{subfigure}
\begin{subfigure}[b]{0.3\textwidth}
	\includegraphics[width=1\textwidth]{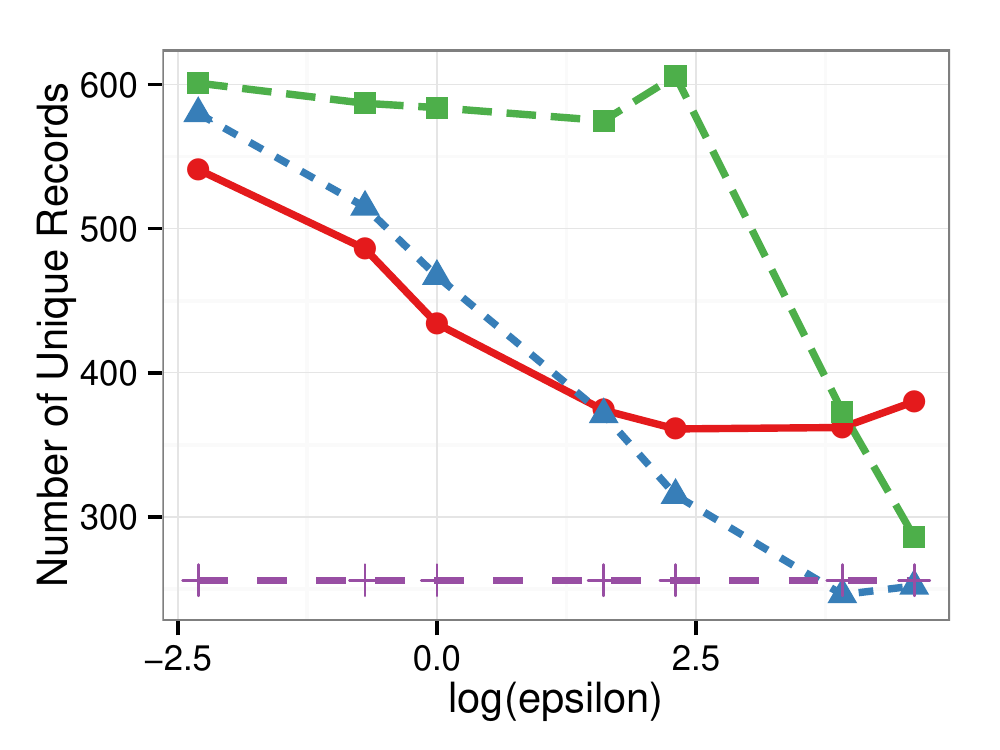}
	\caption{Population uniqueness}
\end{subfigure}
\end{center}
\caption{Simulated attack scenarios on the MDC and charge variables (left and center), and population uniqueness (right).}\label{fig:attacks}
\end{figure}

In this paper, we measure the disclosure risks from two different angles: recoverability of feature values and population uniqueness.
First, we examine whether it is possible to infer the values of sensitive feature given demographic information.
Specifically, if the intruder knows someone's \texttt{age}, \texttt{sex}, \texttt{los}, and \texttt{zip}, we would like to measure the likelihood of getting the correct values as follows:
\begin{align*}
&\text{E}[ \mathbbm{1} ( \text{inferred MDC} \neq \text{correct MDC}) \mid \text{age}, \text{sex}, \text{zip}]\\
&\text{E}[  | \text{inferred charge} - \text{correct charge} | \mid \text{age}, \text{los}, \text{zip}]
\end{align*}
where the inferred values are (1) the most frequent MDC categories and  (2) sample means from conditioned synthetic samples. 
We also measure the population uniqueness based on age, sex, and zip code information.
Figure~\ref{fig:attacks} shows the results from this simulated intruder experiment.
Private records are more difficult to reconstruct if misclassification rates and absolute errors are high. 
The probability of recovering MDC is significantly lower than using a simple bootstrap method, but no one method is distinctly better than the other.
The absolute distance of hospital charges shows that synthetic data has comparable predictive power with the bootstrap method.
Noticeably, the absolute errors are higher when the differential privacy parameters are low, and this finding partially supports our use of differential privacy as a disclosure risk measure.
As can be seen in Figure~\ref{fig:attacks} (right), the perturbed synthetic datasets have more unique samples. 
This is the most distinct characteristics of PeGS compared to other statistical disclosure techniques.
Privacy preserving algorithms, such as $k$-anonymity and $l$-diversity, try to reduce population uniqueness, while PeGS increases the diversity of samples.
The former algorithms apply privacy-preserving transforms on the original data, while the latter algorithm synthesizes a diversified dataset.

\section{Concluding Remarks}\label{sec:conclusion}

In this paper, we proposed a categorical data synthesizer that guarantees prescribed differential privacy or $l$-diversity levels.
The use of a hash function allows the Perturbed Gibbs Sampler to handle high-dimensional categorical data.
The non-parametric modeling of categorial data provides a flexible alternative to traditional (GLM-based) Multiple Imputation techniques.
Additionally, this simple representation of conditional distributions is a crucial component of our block sampling algorithm, which enhances the utility of synthetic data given a fixed privacy budget.

The California Patient Discharge dataset was used to demonstrate the analytical validity and utility of the proposed synthetic methodologies.
Marginal and conditional distributions, as well as regression coefficients of predictive models learned from the synthesized data were compared to those from the original data to quantify the amount of distortion introduced by the synthesization process.
Simulated intruder scenarios were studied to show the confidentiality of the synthesized data.
The empirical studies showed that the proposed mechanisms can provide useful risk-calibrated synthetic data.

Currently, PeGS only deals with categorical variables.
Numeric variables need to be binned to form categorical variables.
Although this approach may be good enough for some applications, brute-force binning ignores numeric similarity or ordering information.
For example, two consecutive values from an ordinal variable are more similar than separated values. 
Consider a size variable with three values: small, medium, and large. 
The ordering information states that similarity(small, medium) $>$ similarity(small, big), but this information is lost if we bin the size variable into three (non-ordered) categories. 
Such semantic correlation cannot be captured in the current synthetic and perturbation model.

In addition to the perturbation step, the hashing step of PeGS also provides some degrees of privacy protection, although it was originally designed for computational efficiency.
When building the PeGS statistical building blocks, each row $\mathbf{x}$ of the original data is hashed based on $h(\mathbf{x}_{-i})$, and  aggregated with other rows with the same hash key, $\{ \mathbf{z} \mid h(\mathbf{z}_{-i})=h(\mathbf{x}_{-i})\}$. 
Although, in this paper, the privacy guarantee of PeGS is analyzed from the perturbation perspective, this aggregation (or hashing) step should be also incorporated for a tighter guarantee of privacy.
The privacy guarantee of PeGS would be affected by different hash resolutions and mechanisms, and this topic needs to be covered in future work.

Although the proposed algorithms show better performance on $\epsilon$-differential privacy and $l$-diversity\footnote{Experimental results on $l$-diversified synthetic data are presented in \citep{Park2013}.} measures, they were only marginally better than the perturbed multiple imputation in other probabilistic disclosure risk measures.
The differential privacy measure may be too conservative for real data, and the probabilistic measure may not exhaustively capture all the attack scenarios.
This is why we provided multiple risk measures.
The connection between the differential privacy and disclosure risks should be further addressed to better evaluate the validity and utility of the synthetic data.  

In practice, multiple disclosure techniques are sequentially mixed to achieve better protection of the records.
For example, PeGS can be applied on top of feature generalization or masking techniques.
Furthermore, some features can be modeled using generalized linear models; for example, numeric features.
It would be worthwhile to investigate cocktails of different statistical disclosure limitation techniques.

\section*{Acknowledgment}
We thank Mallikarjun Shankar for helpful discussions.

\bibliographystyle{abbrvnat}
\bibliography{pegs}

\begin{thebibliography}{43}
\providecommand{\natexlab}[1]{#1}
\providecommand{\url}[1]{\texttt{#1}}
\expandafter\ifx\csname urlstyle\endcsname\relax
  \providecommand{\doi}[1]{doi: #1}\else
  \providecommand{\doi}{doi: \begingroup \urlstyle{rm}\Url}\fi

\bibitem[Abowd and Vilhuber(2008)]{Abowd2008}
J.~M. Abowd and L.~Vilhuber.
\newblock How protective are synthetic data?
\newblock \emph{Privacy in Statistical Databases}, 5262:\penalty0 239--246,
  2008.

\bibitem[Abowd and Woodcock(2001)]{Abowd2001}
J.~M. Abowd and S.~D. Woodcock.
\newblock Disclosure limitation in longitudinal linked data.
\newblock \emph{Confidentiality Diclosure and Data Access: Theory and Practical
  Applications for Statistical Agencies}, pages 215--277, 2001.

\bibitem[Barak et~al.(2007)Barak, Chaudhuri, Dwork, Kale, McSherry, and
  Talwar]{Barak2007}
B.~Barak, K.~Chaudhuri, C.~Dwork, S.~Kale, F.~McSherry, and K.~Talwar.
\newblock Privacy, accuracy, and consistency too: A holistic solution to
  contingency table release.
\newblock In \emph{The 26th ACM SIGMOD-SIGACT-SIGART Symposium on Principles of
  Database Systems}, 2007.

\bibitem[Caiola and Reiter(2010)]{Caiola2010}
G.~Caiola and J.~P. Reiter.
\newblock {Random Forests for Generating Partially Synthetic, Categorical
  Data}.
\newblock \emph{Transactions on Data Privacy}, 3:\penalty0 27--42, 2010.

\bibitem[{Centers for Medicare and Medicaid Services}(2013)]{web:cms}
{Centers for Medicare and Medicaid Services}.
\newblock {Medicare Claims Synthetic Public Use Files (SynPUFs)}.
\newblock
  \url{http://www.cms.gov/Research-Statistics-Data-and-Systems/\\Statistics-Trends-and-Reports/SynPUFs/},
  2013.

\bibitem[Charest(2012)]{Charest2012}
A.-S. Charest.
\newblock Empirical evaluation of statistical inference from
  differentially-private contingency tables.
\newblock \emph{Privacy in Statistical Databases}, 7556:\penalty0 257--272,
  2012.

\bibitem[Clifton and Tassa(2013)]{Clifton2013}
C.~Clifton and T.~Tassa.
\newblock On syntactic anonymity and differential privacy.
\newblock \emph{Transactions on Data Privacy}, 6\penalty0 (2):\penalty0
  161--183, 2013.

\bibitem[Dale and Elliot(2001)]{Dale2001}
A.~Dale and M.~Elliot.
\newblock Proposals for 2001 samples of anonymized records: an assessment of
  disclosure risk.
\newblock \emph{Journal of Royal Statistical Society: Series A}, 2001.

\bibitem[Dalenius and Reiss(1978)]{Dalenius1978}
T.~Dalenius and S.~P. Reiss.
\newblock Data-swapping: A technique for disclosure control (exteded abstract).
\newblock In \emph{Proceedings of the Section on Survey Research Methods},
  1978.

\bibitem[Dankar and Emam(2012)]{Dankar2012}
F.~K. Dankar and K.~E. Emam.
\newblock The application of differential privacy to health data.
\newblock In \emph{Proceedings of Privacy and Anonymity in the Information
  Society (PAIS)}, 2012.

\bibitem[Drechsler and Reiter(2010)]{Drechsler2010}
J.~Drechsler and J.~P. Reiter.
\newblock Sampling with synthesis: A new approach for releasing public use
  census microdata.
\newblock \emph{Journal of the American Statistical Association}, 105\penalty0
  (492):\penalty0 1347--1357, 2010.

\bibitem[Duncan and Lambert(1986)]{Duncan1986}
G.~T. Duncan and D.~Lambert.
\newblock Disclosure-limited data dissemination.
\newblock \emph{Journal of the American Statistical Association}, 1986.

\bibitem[Duncan et~al.(2001)Duncan, Keller-McNulty, and Stokes]{Duncan2001}
G.~T. Duncan, S.~A. Keller-McNulty, and S.~L. Stokes.
\newblock Disclosure risk vs. data utility: The {R-U} confidentiality map.
\newblock Technical report, National Institute of Statistical Sciences, 2001.

\bibitem[Dwork(2006)]{Dwork2006:icalp}
C.~Dwork.
\newblock Differential privacy.
\newblock In \emph{{Proceedings of the 33rd International Colloquium on
  Automata, Languages and Programming}}, volume 4052, pages 1--12, 2006.

\bibitem[Fienberg and McIntyre(2005)]{Fienberg2005}
S.~E. Fienberg and J.~McIntyre.
\newblock Data swapping: Variations on a theme by {Dalenius} and {Reiss}.
\newblock \emph{Journal of Official Statistics}, 21:\penalty0 309--323, 2005.

\bibitem[Fienberg et~al.(2010)Fienberg, Rinaldo, and Yang]{Fienberg2010}
S.~E. Fienberg, A.~Rinaldo, and X.~Yang.
\newblock Differential privacy and the risk-utility tradeoff for
  multi-dimensional contingency tables.
\newblock In \emph{The 2010 International Conference on Privacy in Statistical
  Databases}, pages 187--199, 2010.

\bibitem[Franconi and Stander(2002)]{Franconi2002}
L.~Franconi and J.~Stander.
\newblock A model-based method for disclosure limitation of business microdata.
\newblock \emph{Journal of the Royal Statistical Society, Series D},
  51\penalty0 (1):\penalty0 51--61, 2002.

\bibitem[Friedman et~al.(2010)Friedman, Hastie, and Tibshirani]{Friedman2010}
J.~Friedman, T.~Hastie, and R.~Tibshirani.
\newblock Regularization paths for generalized linear models via coordinate
  descent.
\newblock \emph{Journal of Statistical Software}, 2010.

\bibitem[Fuller(1993)]{Fuller1993}
W.~A. Fuller.
\newblock Masking procedures for microdata disclosure limitation.
\newblock \emph{Journal of Official Statistics}, 9\penalty0 (2):\penalty0
  383--406, 1993.

\bibitem[Gionis et~al.(1999)Gionis, Indyk, and Motwani]{Gionis1999}
A.~Gionis, P.~Indyk, and R.~Motwani.
\newblock Similarity search in high dimensions via hashing.
\newblock In \emph{Proceedings of the 25th Very Large Database}, 1999.

\bibitem[Indyk and Motwani(1998)]{Indyk1998}
P.~Indyk and R.~Motwani.
\newblock Approximate nearest neighbors: Towards removing the curse of
  dimensionality.
\newblock In \emph{Proceedings of 30th Symposium on Theory of Computing}, 1998.

\bibitem[Machanavajjhala et~al.(2007)Machanavajjhala, Kifer, Gehrke, and
  Venkitasubramanian]{Machanavajjhala2007}
A.~Machanavajjhala, D.~Kifer, J.~Gehrke, and M.~Venkitasubramanian.
\newblock $l$-diversity: Privacy beyond $k$-anonymity.
\newblock \emph{Transactions on Knowledge Discovery from Data}, 1, 2007.

\bibitem[Machanavajjhala et~al.(2008)Machanavajjhala, Kifer, Abowd, Gehrke, and
  Vilhuber]{Machanavajjhala2008}
A.~Machanavajjhala, D.~Kifer, J.~Abowd, J.~Gehrke, and L.~Vilhuber.
\newblock {Privacy: Theory meets Practice on the Map}.
\newblock In \emph{{Proceedings of the 24th International Conference on Data
  Engineering}}, 2008.

\bibitem[McClure and Reiter(2012)]{McClure2012}
D.~McClure and J.~P. Reiter.
\newblock Differential privacy and statistical disclosure risk measures: An
  investigation with binary synthetic data.
\newblock \emph{Transactions on Data Privacy}, 5:\penalty0 535--552, 2012.

\bibitem[McSherry and Talwar(2007)]{McSherry2007}
F.~McSherry and K.~Talwar.
\newblock Mechanism design via differential privacy.
\newblock In \emph{Proceedings of the 48th Annual Symposium of Foundations of
  Computer Science}, 2007.

\bibitem[Park et~al.(2013)Park, Ghosh, and Shankar]{Park2013}
Y.~Park, J.~Ghosh, and M.~Shankar.
\newblock Lugs: A scalable non-parametric data synthesizer for privacy
  preserving big health data publication.
\newblock In \emph{International Conference on Machine Learning WHEALTH 2013},
  2013.

\bibitem[Polettini(2003)]{Polettini2003}
S.~Polettini.
\newblock Maximum entropy simulation for microdata protection.
\newblock \emph{Statistics and Computing}, 13:\penalty0 307--320, 2003.

\bibitem[Raghunathan et~al.(2003)Raghunathan, Reiter, and
  Rubin]{Raghunathan2003}
T.~E. Raghunathan, J.~P. Reiter, and D.~B. Rubin.
\newblock Multiple imputation for statistical disclosure limitation.
\newblock \emph{Journal of Official Statistics}, 19\penalty0 (1):\penalty0
  1--16, 2003.

\bibitem[Reiter(2003)]{Reiter2003}
J.~P. Reiter.
\newblock Using {CART} to generate partially synthetic, public use microdata.
\newblock \emph{Journal of Official Statistics}, \penalty0 (441-462), 2003.

\bibitem[Reiter(2005{\natexlab{a}})]{Reiter2005}
J.~P. Reiter.
\newblock Releasing multiply imputed, synthetic public use microdata: an
  illustration and empirical study.
\newblock \emph{Journal of the Royal Statistical Society, Series A},
  168:\penalty0 185--205, 2005{\natexlab{a}}.

\bibitem[Reiter(2005{\natexlab{b}})]{Reiter2005b}
J.~P. Reiter.
\newblock Estimating risks of identification disclosure in microdata.
\newblock \emph{Journal of American Statistical Association}, 100\penalty0
  (472):\penalty0 1103--1112, 2005{\natexlab{b}}.

\bibitem[Reiter and Drechsler(2010)]{Reiter2010}
J.~P. Reiter and J.~Drechsler.
\newblock Releasing multiply imputed synthetic data generated in two stages to
  protect confidentiality.
\newblock \emph{Statistica Sinica}, 20:\penalty0 405--421, 2010.

\bibitem[Rubin(1987)]{Rubin1987}
D.~B. Rubin.
\newblock \emph{Multiple Imputation for Nonresponse in Surveys}.
\newblock Wiley, 1987.

\bibitem[Rubin(1993)]{Rubin1993}
D.~B. Rubin.
\newblock Discussion: Statistical disclosure limitation.
\newblock \emph{Journal of Official Statistics}, 9:\penalty0 462--468, 1993.

\bibitem[Sakshaug and Raghunathan(2011)]{Sakshaug2011}
J.~W. Sakshaug and T.~E. Raghunathan.
\newblock Synthetic data for small area estimation.
\newblock \emph{Privacy in Statistical Databases}, 6344:\penalty0 162--173,
  2011.

\bibitem[Soria-Cormas and Drechsler(2013)]{SoriaCormas2013}
J.~Soria-Cormas and J.~Drechsler.
\newblock Evaluating the potential of differential privacy mechanisms for
  census data.
\newblock In \emph{UNECE Conference of European Statisticans}, 2013.

\bibitem[Sweeney(2002)]{Sweeney2002}
L.~Sweeney.
\newblock k-anonymity: a model for protecting privacy.
\newblock \emph{Int. J. Uncertain. Fuzziness Knowl.-Based Syst.}, 10:\penalty0
  557--570, October 2002.
\newblock ISSN 0218-4885.

\bibitem[Vreeken and Siebes(2008)]{Vreeken2008}
J.~Vreeken and A.~Siebes.
\newblock Filling in the blanks - {KRIMP} minimisation for missing data.
\newblock In \emph{the 8th IEEE International Conference on Data Mining}, pages
  1067--1072, 2008.

\bibitem[Weinberger et~al.(2009)Weinberger, Dasgupta, Langford, Smola, and
  Attenberg]{Weinberger2009}
K.~Weinberger, A.~Dasgupta, J.~Langford, A.~Smola, and J.~Attenberg.
\newblock Feature hashing for large scale multitask learning.
\newblock In \emph{{Proceedings of the 26th International Conference on Machine
  Learning}}, 2009.

\bibitem[Willenborg and de~Waal(2001)]{Willenborg2001}
L.~Willenborg and T.~de~Waal.
\newblock \emph{Elements of Statistical Disclosure Control}, volume 155.
\newblock Springer, 2001.

\bibitem[Winkler(2003)]{Winkler2003}
W.~E. Winkler.
\newblock A contingency-table model for imputing data satisfying analytic
  constraints.
\newblock \emph{U.S. Census Bureau Statistical Research Division Research
  Report Series}, 2003.

\bibitem[Winkler(2010)]{Winkler2010}
W.~E. Winkler.
\newblock General discrete-data modeling methods for producing synthetic data
  with reduced re-identification risk that preserve analytic properties.
\newblock \emph{U.S. Census Bureau Statistical Research Division Research
  Report Series}, 2010.

\bibitem[Yang et~al.(2012)Yang, Fienberg, and Rinaldo]{Yang2012}
X.~Yang, S.~E. Fienberg, and A.~Rinaldo.
\newblock Differential privacy for protecting multi-dimensional contingency
  table data: Extensions and applications.
\newblock \emph{Journal of Privacy and Confidentiality}, 4\penalty0
  (1):\penalty0 101--125, 2012.

\end{thebibliography}

\end{document}